\documentclass[sigconf]{acmart} 
\AtBeginDocument{%
  }

\copyrightyear{2025} 
\acmYear{2025} 
\setcopyright{cc}
\setcctype{by-nc}
\acmConference[McGE '25]{Proceedings of the 3rd International Workshop on
Multimedia Content Generation and Evaluation: New Methods and
Practice}{October 27--28, 2025}{Dublin, Ireland}
\acmBooktitle{Proceedings of the 3rd International Workshop on Multimedia
Content Generation and Evaluation: New Methods and Practice (McGE '25),
October 27--28, 2025, Dublin, Ireland}\acmDOI{10.1145/3746278.3759376}
\acmISBN{979-8-4007-2060-4/2025/10}

\usepackage{multirow}

\usepackage{pifont}
\newcommand{\cmark}{\ding{51}}%
\newcommand{\xmark}{\ding{55}}%

\begin{document}

\title{Towards a General-Purpose Zero-Shot Synthetic Low-Light Image and Video Pipeline}

\author{Joanne Lin}
\authornote{Both authors contributed equally to this research.}
\authornote{Corresponding author.}
\email{joanne.lin@bristol.ac.uk}
\orcid{0009-0000-4396-3550}
\affiliation{%
  \institution{University of Bristol}
  \city{Bristol}
  \country{United Kingdom}
}

\author{Crispian Morris}
\authornotemark[1]
\email{crispian.morris@bristol.ac.uk}
\affiliation{%
  \institution{University of Bristol}
  \city{Bristol}
  \country{United Kingdom}
}

\author{Ruirui Lin}
\email{r.lin@bristol.ac.uk}
\affiliation{%
  \institution{University of Bristol}
  \city{Bristol}
  \country{United Kingdom}
}

\author{Fan Zhang}
\email{fan.zhang@bristol.ac.uk}
\affiliation{%
  \institution{University of Bristol}
  \city{Bristol}
  \country{United Kingdom}
}

\author{David Bull}
\email{dave.bull@bristol.ac.uk}
\affiliation{%
  \institution{University of Bristol}
  \city{Bristol}
  \country{United Kingdom}
}

\author{Nantheera Anantrasirichai}
\email{N.Anantrasirichai@bristol.ac.uk}
\affiliation{%
  \institution{University of Bristol}
  \city{Bristol}
  \country{United Kingdom}
}

\renewcommand{\shortauthors}{Lin et al.}

\begin{abstract}
Low-light conditions pose significant challenges for both human and machine annotation. This in turn has led to a lack of research into machine understanding for low-light images and (in particular) videos. A common approach is to synthesize low-light datasets from existing datasets with high-quality annotations. However, these approaches are often limited through the use of unrealistic noise models, or require metadata information often unavailable in public datasets. In this paper, we propose a new Degradation Estimation Network (DEN), which synthetically generates realistic standard RGB (sRGB) noise without the requirement for camera metadata. This is achieved by estimating the parameters of physics-informed noise distributions, trained in a self-supervised manner. This zero-shot approach allows our method to generate synthetic noisy content with a diverse range of realistic noise characteristics, unlike other methods which focus on recreating the noise characteristics of the training data. We evaluate our proposed synthetic pipeline using various methods trained on its synthetic data for typical low-light tasks including synthetic noise replication, video enhancement, and object detection, showing improvements of up to 24\% KLD, 21\% LPIPS, and 62\% AP$_{50-95}$, respectively.
\end{abstract}





\maketitle


\section{Introduction}

\begin{figure}[!t]
 \centering
 \includegraphics[width=0.85\linewidth]{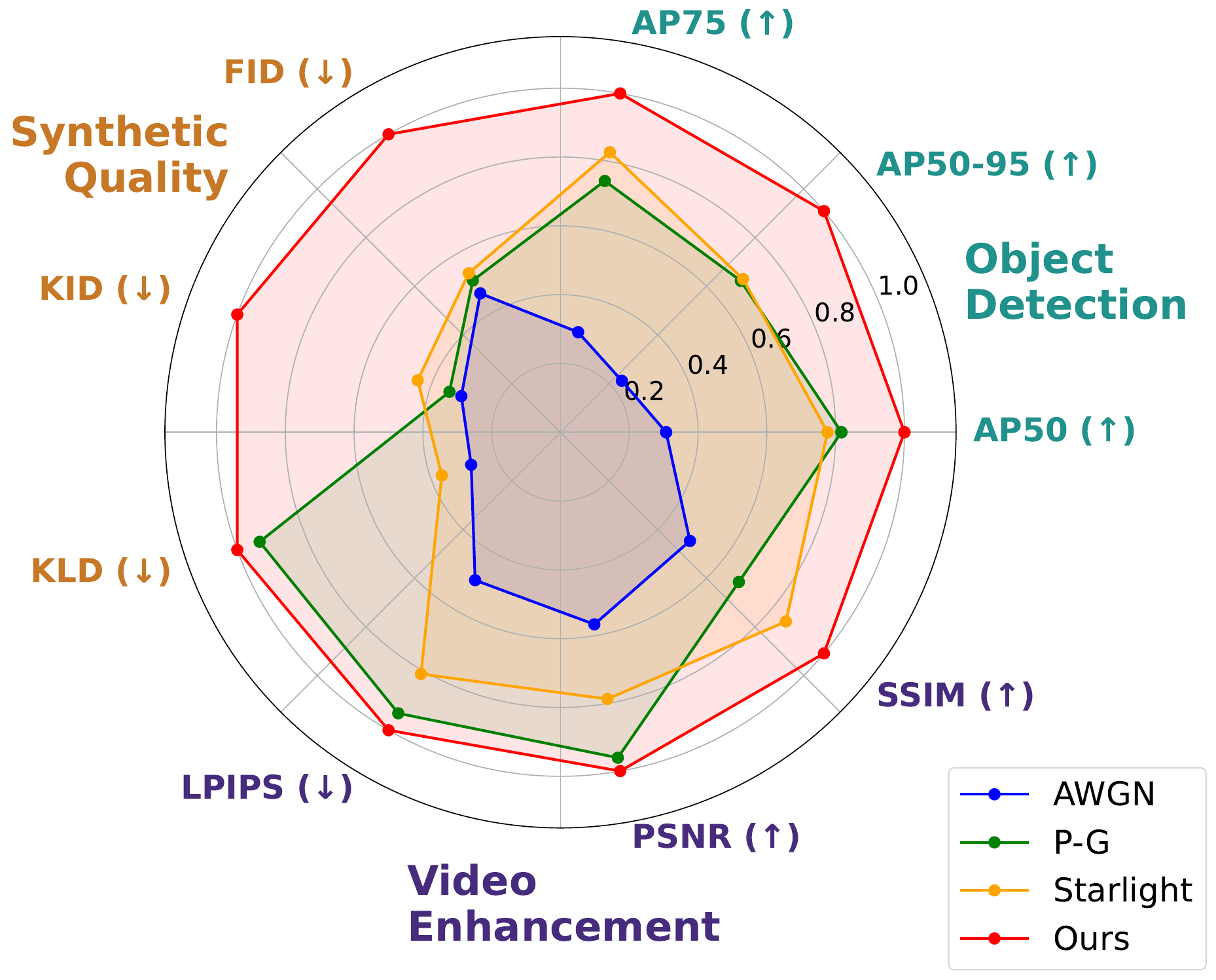}
 \caption{Comparison of our proposed method with existing synthetic noise pipelines, where our model outperforms these methods across numerous benchmarks. Results are taken from Table \ref{tab:results} and normalized for visualization. For metrics where lower is better, indicated by $\downarrow$, the results are inverted.}
 \Description{Radar plot using several metrics where our model outperforms the synthetic noise pipelines.}
 \label{fig:radar_plot}
\end{figure}

The performance of computer vision methods for various image and video processing tasks (e.g. classification, action recognition, object detection, object tracking, instance segmentation, etc.) has advanced significantly in recent years. This is due primarily to advances in deep learning methods and increased computational power, allowing deeper and more complex networks to be trained. The demand for video-based research is evidenced by the numerous international challenges in recent years ~\cite{sigurdsson2016charades, Perazzi2016davis, xu2018ytvos, Damen2018EPICKITCHENS, yang2019ytvis, huang2021got10k, anantrasirichai2025AI}. This has, in turn, enabled many practical applications including autonomous driving, digital surveillance, augmented/virtual reality (AR/VR), media production, and entertainment.

Despite these dramatic advances, research on processing videos captured under low-light conditions remains underdeveloped. Low-light acquisition introduces numerous degradations such as low-contrast and motion blur, alongside sensor noise. All of these impact the performance of algorithms trained on normal-light data~\cite{Yi2024trackinglowlight}. A further challenge is the lack of paired data, which is difficult to annotate accurately under low-light conditions without clear ground truth. As a result, only a limited number of datasets are available that offer adequate annotations for low-light video perception tasks~\cite{ye2022darktrack, Li2024LLEVOS, wang2024lmot, liu2024ntvot211}. 

To overcome this, a common approach is to first apply low-light video enhancement (LLVE) methods~\cite{guo2020zerodce, wang2021sdsd, fu2023did, lin2024stasunet} before performing the desired downstream task. This enables models trained on normal-light conditions to process the video without having been trained specifically on low-light. Low-light enhancement is, however, an ill-posed problem, where a single low-light input could be mapped to multiple potential outputs. This may introduce greater variability into the system, making it more difficult to solve a downstream task with a single solution. Moreover, adding extra pre-processing steps increases computational complexity, especially for deep-learning-based methods, thereby limiting real-time applications. In addition, imperfect pre-processing of input data may produce artifacts that deteriorate the performance of downstream models
~\cite{lin2025segmlowlight, chen2025qualityandperception}.

An alternative approach is to generate synthetic data for training and evaluating the models. Using synthetic pipelines removes the need to source suitable low-light datasets with the appropriate annotations; an existing clean dataset with annotations can be used instead. Although there are numerous pipelines available for creating synthetic low-light images and videos, each comes with limitations. Traditional pipelines often employ physics-informed distributions, which require manual selection of appropriate noise parameters~\cite{foi2008noiseapprox, anantrasirichai2015cwttexture} and fail to replicate sRGB-specific noise characteristics arising from in-camera processing pipelines while more complex pipelines may be tailored to noise types from specific datasets, which significantly restricts their applications, or they may require camera-specific information that is not available in most publicly accessible datasets for downstream tasks.

In this context, we introduce a novel, general-purpose, zero-shot framework to address the limitations present in existing synthetic data pipelines. Our approach is capable of capturing a diverse range of realistic degradation types, by estimating a vector of noise parameters which is used to sample noise maps from the corresponding physics-informed distributions. The main contributions are summarized as follows:

\begin{itemize}
    \item The \textbf{first general-purpose zero-shot synthetic low-light pipeline} that outputs realistic low-light sRGB images and videos with the same noise characteristics as reference real low-light content. Our approach does not include any real low-light data during training, and allows desired noise characteristics to be applied onto any given input. 
    \item Unlike the only existing zero-shot deep-learning method proposed by~\cite{luo2023similarityminmax}, which is limited to Gaussian-distributed noise, our method offers a broad range of noise types and parameters, making it suitable for real-world applications.
    \item A \textbf{novel Degradation Estimation Network (DEN)} which can predict the parameters of physics-informed distributions to model the noise accurately onto low-light videos. This approach differs from existing works, which typically model output-specific noise based only on their specific training data.
    \item A \textbf{novel self-supervised training strategy} that exposes the DEN to a diverse distribution of noise parameters, enabling it to robustly learn and generalize across a wide range of noise characteristics.
\end{itemize}

The proposed method has been evaluated across various tasks, including synthetic noise replication, video enhancement, and object detection, demonstrating improvements of up to 24\% KLD, 21\% LPIPS, and 62\% AP$_{50-95}$, respectively. Comparative analysis against established noise models, including Additive White Gaussian Noise (AWGN), Poisson-Gaussian noise (P-G)~\cite{foi2008noiseapprox}, and the Starlight noise generator~\cite{monakhova2022starlight}, reveals our method's superior performance, as shown in Fig. \ref{fig:radar_plot}.


\section{Related Work}
\label{sec:works}
\subsection{RAW Noise Pipelines}

RAW image data inherently contains various types of noise, often modeled using statistical distributions that reflect the physical processes of image capture. Many researchers make use of these physical phenomena to synthesize noisy data for training and analysis. The most commonly modeled noise type is read noise, which is typically modeled using a Gaussian distribution. Other forms of noise, such as shot noise, banding noise, or quantization noise, are also represented using appropriate statistical models. While these methods are capable of modeling many noise types using only statistics, they require manual parameter selection to ensure a realistic output. This is evidenced in work by Wei et al.~\cite{wei2021physicsmodeling}, where meticulous experimentation was required to estimate the noise parameters for five different cameras.

More recently, there has been an increase in the use of deep-learning methods to synthesize realistic low-light images, without the need for parameter estimation~\cite{Cao2023ISOdependentnoise, Zhang2023RAWnoise}. These methods are often described as `physics-guided', incorporating physics-informed distributions within the neural networks for generating realistic low-light imagery. Cao et al.~\cite{Cao2023ISOdependentnoise} developed a method that models a variety of different noise types, but it requires the availability of ISO information for the image. Meanwhile, Zhang et al.~\cite{Zhang2023RAWnoise} proposed a more general-purpose pipeline, without the need for camera metadata, but their method is restricted to their training data. Monakhova et al.~\cite{monakhova2022starlight} proposed a generative adversarial network (GAN) for synthesizing noisy RAW videos to train their video denoiser. Similar to other physics-guided pipelines, their generator initially samples from several statistical distributions with learned parameters for their dataset, followed by processing the video through a U-Net~\cite{ronneberge2015unet} to learn other complex noise types. However, like~\cite{Zhang2023RAWnoise}, their method requires re-training when applied to different cameras or datasets, highlighting a common limitation in the generalizability of current noise synthesis approaches.

\subsection{sRGB Noise Pipelines}

Many synthetic pipelines predominantly use the RAW format to model noise based on physics principles. However, sRGB images inherently exhibit additional degradation types which need to be considered. One of the main noise types is spatially-correlated noise, which is often caused by compression algorithms, denoising algorithms and other processes found in the image signal processing (ISP) pipeline. Due to the complex nature of noise types like this, limited research has been conducted for synthetic noisy sRGB pipelines. Deep learning approaches have been explored, but these usually require further information about the image, such as camera parameters and ISO settings. For example, Kousha et al.~\cite{kousha2022sRGBnormflow} synthesized noisy sRGB images by leveraging normalizing flows, a family of generative models which use invertible transformations to translate data from one domain to another. Their method includes a specific linear flow layer conditioned on the camera metadata and gain (ISO) to ensure accurate synthetic noisy images. 

Fu et al.~\cite{fu2023sRGBneighbornoise} introduced an alternative approach for generating noisy sRGB images that accounts for spatially-correlated noise without the need for ISO data of the captured image. They proposed a `gain estimation network' that enables the model to autonomously synthesize noise without additional camera information. However, their method is still limited in terms of applications, as it is unable to generalize across different camera models, requiring re-training for each distinct camera.

\subsection{Low-Light sRGB Pipelines}

Existing pipelines for generating low-light images and videos often rely on simple noise models that primarily capture common noise types such as read and shot noise, and primarily focus on modifying the brightness and contrast of the data. This approach generally lacks the sophistication required to replicate realistic noise conditions effectively.

Lv et al.~\cite{lv2021agllnet} proposed a method for generating low-light images which mimics the ISP pipeline in low-light photography. They first reduce the brightness of the images using a combination of linear and power transformations, before converting the images to a Bayer format. This synthetically creates a `RAW' low-light image, where physics-informed noise types (i.e. read and shot noise) are applied, before converting back into sRGB format via demosaicing techniques. Cui et al.~\cite{cui2021maet} also proposed a similar approach to simulating the ISP pipeline, but factor in further image transformations such as white balance and gamma correction to enhance realism.
Luo et al.~\cite{luo2023similarityminmax} proposed a deep-learning based method for synthesizing low-light images, by training a network to estimate adjustment maps to reduce the brightness of the input images, conditioned by exposure map. However, their method only draws noise maps from Gaussian distributions, and does not consider signal-dependent noise.

For videos, Zhou et al.~\cite{zhou2022lednet} proposed a synthetic low-light pipeline focusing on blur degradations due to low-light conditions. Unlike the simpler brightness adjustment techniques, they use a modified version of Zero-DCE~\cite{guo2020zerodce} to reduce the brightness of the image, instead of simple linear or power transformations. They use a pre-existing method, CycleISP~\cite{zamir2020cycleisp}, for synthesizing noise.


\section{Proposed Method}
\label{sec:methods}

\begin{figure*}[!t]
\centering
    \includegraphics[width=1.04\textwidth]{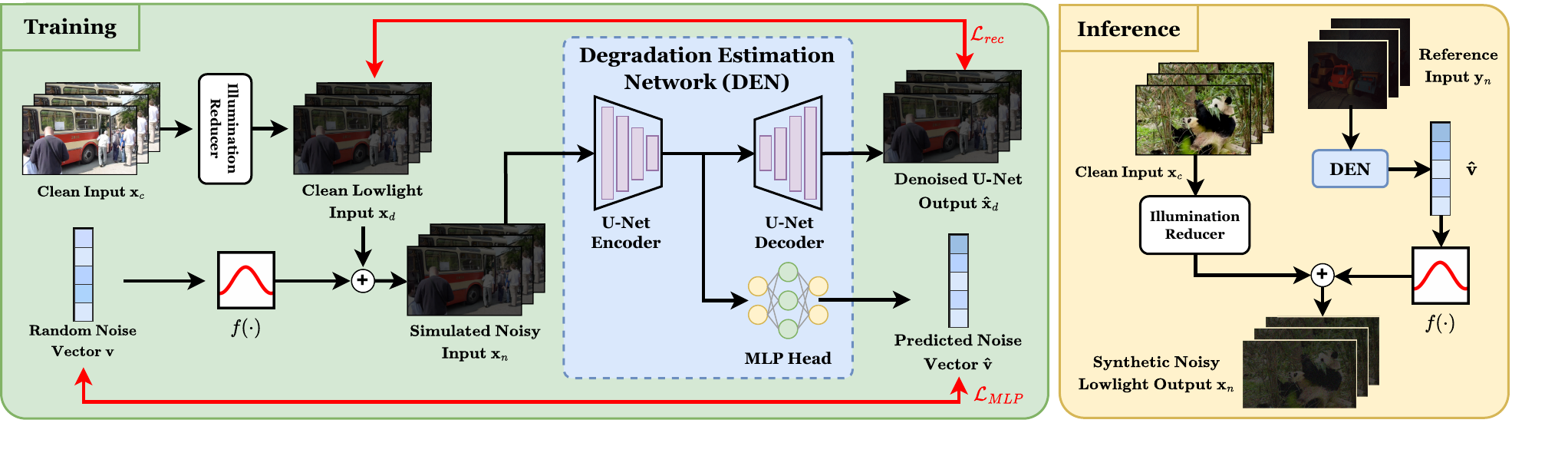}
    \vspace{-8mm}
    \caption{Overview of the proposed framework. The Degradation Estimation Network (DEN) is based on a U-Net~\cite{ronneberge2015unet} with an MLP head attached to the encoder, used to predict a noise vector $\hat{\mathbf{v}}$ with denoising guidance. In training, the illumination reducer and noise simulator function $f(\cdot)$ are used to create simulated noisy inputs for DEN. In inference, a real low-light input $\mathbf{y}_n$ is used to obtain a noise vector to be applied onto the clean input.}
    \vspace{-4mm}
    \label{fig:pipeline}
    \Description{The pipeline of the proposed framework, showing the training and inference steps of using our Degradation Estimation Network (DEN) to predict the noise vector $\hat{\mathbf{v}}$ which is used to recreate the input noise characteristics.}
\end{figure*}

We propose a new zero-shot generic synthetic noise pipeline capable of generating realistic low-light videos that accurately replicates the noise characteristics observed in a given reference video. This pipeline is also applicable to images by simply reducing the temporal dimension to one. To allow for zero-shot capabilities and to ensure robustness across various camera settings and noise characteristics, we employ a self-supervised training strategy that involves the generation of images and videos with a large and representative range of noise characteristics. Our approach guarantees that our method adapts to diverse noise patterns typically seen across different camera systems or the same system with different settings. The training and inference processes are illustrated in Fig. \ref{fig:pipeline}, and Fig. \ref{fig:pipeline}, respectively.

In the \textbf{training phase}, our methodology begins by attenuating the brightness of clean input data using linear and power transformations, referred to as the `illumination reducer' in Fig. \ref{fig:pipeline}. We set
\begin{equation}
 \mathbf{x}_d = \alpha \mathbf{x}_c^{1/\gamma},
 \label{eq:lowlight}
\end{equation}
where $\mathbf{x}_d$ is the darkened image and $\mathbf{x}_c$ is the clean input image. The coefficients $\alpha$ and $\gamma$ are sampled from uniform distributions, empirically $\alpha \sim \mathcal{U}(0.05, 0.3)$ and $\gamma \sim \mathcal{U}(0.1, 1)$, to vary the intensity and gamma correction of the synthesized imagery respectively.

Following this darkening process, random values are selected for each noise type described in Section~\ref{ssec:physics_noise}, and applied to the input data. Subsequently, we pass this video into our Degradation Estimation Network (DEN), which learns to estimate the vector of the parameters, referred to as the noise vector $\mathbf{v}$, from the distributions that the noise map is sampled from. Concurrently, the DEN is trained to perform denoising used for optimization ($\mathcal{L}_{rec}$ in Fig.~\ref{fig:pipeline}), enhancing its ability to generalize across various noise conditions. 

During the \textbf{inference phase}, real low-light images or videos serve as the reference inputs $\mathbf{y}_n$ to the DEN, which predicts the noise vector that most accurately represents the noise characteristics present in the reference video. The predicted noise vector is then used to simulate noise, effectively replicating the noise patterns found in the low-light conditions of the reference data.

The overall inference process can be summarized as
\begin{equation}
\begin{split}
 \mathbf{x}_n &= \mathbf{x}_d + f(\mathbf{\hat{v}}),\\
 \mathbf{\hat{v}} &= \text{DEN}(\mathbf{y}_n),
 \end{split}
 \label{eqn:synthesis}
\end{equation}
where $\mathbf{x}_n$ stands for the synthesized low-light image or video, and $\mathbf{y}_n$ is the reference videos with real noise. $\mathbf{\hat{v}}$ represents the estimated noise vector which contains the parameters for the noise distribution models, and $f(\cdot)$ is the noise simulation function comprised of various noise models. 

Crucially, our method ultimately allows for the training of any model on any reference noise types. This process involves analyzing the input data to constrain the range of values for the random noise vector $\mathbf{v}$ used during inference of our method in the downstream model training. By doing so, the model in question can be specifically trained on targeted noise characteristics or specific downstream tasks, greatly enhancing its effectiveness.

\subsection{Physics-Based Noise}
\label{ssec:physics_noise}

In our framework, we focus on physics-based noise to accurately simulate realistic noise characteristics prevalent in low-light conditions. Five predominant noise types are modeled as follows. 


\paragraph{Heteroscedastic noise.} Read and shot noise are typical in low-light imagery ~\cite{ElGamal2005cmos, boncelet2009noisemodels}. Read noise arises from electrical noise inherent in the signal readout process from the camera sensor. It is a combination of a variety of noise types that can generally be modeled using a Gaussian distribution~\cite{boncelet2009noisemodels}. Shot noise, on the other hand, results from the quantum nature of light. During image capture, some photons may not have reached the sensor, leading to image degradation. This type of noise is particularly more noticeable in low-light scenes where photon counts are low. This effect can be modeled using a Poisson distribution. To simulate both types of noise, we employ the heteroscedastic noise model proposed by Foi et al.~\cite{foi2008noiseapprox}, which approximates the Poisson distribution as zero-mean Gaussian. The heteroscedastic noise model is beneficial as it can approximate shot noise, a type of signal-dependent noise that is not inherently additive, to be treated in an additive manner. The heteroscedastic noise model is mathematically described for a video frame of height $H$, width $W$ and channels $C$ as
\begin{equation}
 \mathbf{n}_h \sim \mathcal{N}(0, \sigma_s^2\mathbf{x}+\sigma_r^2), \quad \mathbf{n}_h \in \mathbb{R}^{C \times H \times W},
 \label{eqn:noise_heteroscedastic}
\end{equation}
where $\mathbf{n}_h$, $\mathbf{x}$, $\sigma_s^2$, and $\sigma_r^2$ are the heteroscedastic noise, the clean image signal, the signal-dependent (shot noise) variance coefficient, and the signal-independent (read noise) variance, respectively.

\paragraph{Quantization noise.} Quantization noise arises during the process of converting the analog signal (with continuous values) into a digital signal (with discrete values). This conversion often introduces visual artifacts in the form of grainy or blocky images, which is especially noticeable in images with low bit depth. We model this using a uniform distribution 
\begin{equation}
 \mathbf{n}_q \sim \mathcal{U}(0, \lambda_q), \quad \mathbf{n}_q \in \mathbb{R}^{C \times H \times W},
 \label{eqn:noise_quantization}
\end{equation}
where $\mathbf{n}_q$ is the quantization noise map, and $\lambda_q$ is the upper bound for the quantization noise interval.

\paragraph{Banding noise.}  Banding noise refers to the horizontal or vertical lines that become prominent at high ISO, commonly used in low-light photography and videography to capture content without the need for long shutter speeds (which causes motion blur). This type of noise is highly camera-specific. Banding noise, like vertical banding as outlined by Monakhova et al.~\cite{monakhova2022starlight}, is modeled using a Gaussian distribution centered at zero, with the variability of band positions encapsulated by the standard deviation $\sigma_b$. The noise model is represented as 

\begin{equation}
 \mathbf{n}_b \sim \mathcal{N}(0, \sigma_b^2), \quad \mathbf{n}_b \in \mathbb{R}^{C \times 1 \times W},
 \label{eqn:noise_banding}
\end{equation}
where $\mathbf{n}_b$ is the banding noise map. Additionally, we include temporal banding noise $\mathbf{n}_{b_t}$ (as described in~\cite{monakhova2022starlight}), where banding noise maintains consistency throughout the video, characterized by a standard deviation of $\sigma_{b_t}$ to model noise persistence over time.

\paragraph{Periodic noise.} Periodic noise appears as a repeating pattern, often caused by electrical interference during the image capturing process. It is characterized by regularly spaced artifacts such as stripes or grids. Periodic noise is distinct from banding noise due to its strictly periodic nature. Unlike random noise, periodic noise can be effectively identified in the frequency domain, where it exhibits as distinct peaks. We follow the periodic noise implementation described in~\cite{monakhova2022starlight}, which is formulated by

\begin{equation}
 \mathbf{n}_p[r, c] = 
 \begin{cases} 
 \mathcal{F}^{-1}(\mathbf{n}_{p_1}), & \text{if} \:\: c = 0, \\
 \mathcal{F}^{-1}(\mathbf{n}_{p_2} + \mathbf{n}_{p_3}i), & \text{if} \:\: c = W/4, \\
 \mathcal{F}^{-1}(\mathbf{n}_{p_2} - \mathbf{n}_{p_3}i), & \text{if} \:\: c = 3W/4, \\
 0, & \text{otherwise},
 \end{cases}
 \label{eqn:noise_periodic}
\end{equation}
where $r$ and $c$ index the row and column of the image, respectively, and $\mathcal{F}^{-1}$ is the inverse Fourier transform. $\mathbf{n}_{p_1}$, $\mathbf{n}_{p_2}$ and $\mathbf{n}_{p_3}$ are random variables, each sampled from zero-mean Gaussian distributions with standard deviations $\sigma_{p_1}$, $\sigma_{p_2}$ and $\sigma_{p_3}$ respectively. 

\subsection{Degradation Estimation Network (DEN)}

Our proposed DEN is designed to replicate the noise characteristics of a given reference low-light image or video $\mathbf{y}_n$. In practice, it is trained to predict a noise vector $\mathbf{\hat{v}}$ that configures the noise simulator $f(\cdot)$ with the appropriate parameters.
Referring to Section \ref{ssec:physics_noise}, the DEN aims to predict the following parameters: $\sigma_s$, $\sigma_r$, $\lambda_q$, $\sigma_b$, $\sigma_{b_t}$, $\sigma_{p_1}$, $\sigma_{p_2}$ and $\sigma_{p_3}$.

The architecture of the DEN is built upon a U-Net framework~\cite{ronneberge2015unet}, with an additional multi-layer perceptron (MLP)~\cite{rosenblatt1958perceptron} head following the encoder. The MLP head is tasked with estimating the noise vector for simulating the noise in the reference data, whereas the decoder reconstructs the denoised version of the input video. The reasoning behind this design is to enable the encoder to effectively discriminate noise from the underlying image content, guiding the MLP head to precise noise vector predictions.

The DEN loss function comprises two components, $\mathcal{L}_{MLP}$ and $\mathcal{L}_{rec}$, to train the weights in both heads of the network. The loss function for the MLP head, $\mathcal{L}_{MLP}$, is calculated using the mean-squared error (MSE) between the predicted noise vector $\mathbf{\hat{v}}$ and the target noise vector $\mathbf{v}$, defined as 
\begin{equation}
 \mathcal{L}_{MLP} = \frac{1}{n} \sum_{i=1}^{n} (v_i - \hat{v}_i)^2,
\end{equation}
where $n$ is the number of elements in vector $\mathbf{v}$, $v_i$ is the $i^\text{th}$ element in the input noise vector, and likewise, $\mathbf{\hat{v}}_i$ is the $i^\text{th}$ element in the predicted noise vector.

The second component, $L_{rec}$, is a reconstruction loss to train the decoder head to denoise the input, defined as
\begin{equation}
 \mathcal{L}_{rec} = \frac{1}{n} \sum_{i=1}^{n} \lvert \mathbf{x}_d - \mathbf{\hat{x}}_d \rvert,
\end{equation}
where $\mathbf{x}_d$ is the darkened clean input, and $\mathbf{\hat{x}_d}$ is the reconstructed denoised input.
Adding the two losses together, we have
\begin{equation}
 \mathcal{L}_{DEN} = \lambda_1 \mathcal{L}_{MLP} +
 \lambda_2 \mathcal{L}_{rec},
\end{equation}
where $\lambda_1$ and $\lambda_2$ are the adjustable hyperparameters for each loss.

The denoising head employs the $\ell_1$ loss function due to its robustness to outliers, enhancing the stability of denoising operations. Conversely, for estimating the noise vector $\mathbf{v}$, we utilize the $\ell_2$ loss function, which penalizes larger deviations more heavily, ensuring that the estimated noise closely approximates actual noise characteristics observed in the data.

\subsection{Noise Simulator $f(\cdot)$}
\label{ssec:noise_simulator}

The noise simulator function $f(\cdot)$ is designed to synthesize realistic noise maps present in images and videos by leveraging the physics-based noise models detailed in Section \ref{ssec:physics_noise}. This function takes a noise parameter vector $\mathbf{v}$, predicted by the DEN using the real noisy input. The vector $\mathbf{v}$ compiles the necessary parameters to control the intensity and characteristics of each noise type: $\mathbf{v} = [\sigma_s, \sigma_r, \lambda_q, \sigma_b, \sigma_{b_t}, \sigma_{p_1}, \sigma_{p_2}, \sigma_{p_3}]$.

The final simulated noise map is defined as the summation of all individual noise components, expressed as

\begin{equation}
 f(\mathbf{v}) = \mathbf{n}_h + \mathbf{n}_q + \mathbf{n}_b + \mathbf{n}_{b_t} + \mathbf{n}_p,
 \label{eqn:noise_simulator_f}
\end{equation}
where $\mathbf{n}_h, \mathbf{n}_q, \mathbf{n}_b, \mathbf{n}_{b_t}$ and $\mathbf{n}_p$ represent the specific noise maps generated by sampling from their respective distributions or applying their defined procedures (as detailed in Equations \ref{eqn:noise_heteroscedastic}-\ref{eqn:noise_periodic}) using the parameters contained within the input vector $\mathbf{v}$.


\section{Experimental Results and Discussion}
\label{sec:results}

\subsection{Implementation Details}
\label{subsec:deets}

We utilize the YouTube-VOS dataset~\cite{xu2018ytvos} for training our network, which includes 3,471 videos representative of diverse real-world scenarios. It also features extensive variability in both object types and movement dynamics for greater generalizability. We train our method for 10 epochs with a batch size of 2 and 16 frames per video. We employ the Adam optimizer with a learning rate of 0.0002, $\beta_1$ of 0.5 and $\beta_2$ of 0.999. Weights $\lambda_1$ and $\lambda_2$ are both set to 1.

For performance benchmarking, we compare our framework against several noise models, including Additive White Gaussian Noise (AWGN), Poisson-Gaussian noise (P-G) as approximated by Foi et al.~\cite{foi2008noiseapprox}, and the Starlight noise generator proposed by Monakhova et al.~\cite{monakhova2022starlight}. The Starlight model is fully re-trained under identical settings as ours and adapted for processing sRGB videos instead of RAW inputs. The consistent training strategy ensures a fair comparison. Importantly, we do not compare our model against other sRGB deep-learning pipelines, such as~\cite{kousha2022sRGBnormflow, fu2023sRGBneighbornoise}, as they require camera metadata, which would otherwise result in an unfair comparison, and this information is unknown in many real-world cases.

\begin{table*}[!t]
 \caption{Quantitative Results of Synthetic Pipelines Across Multiple Experiments. The best performers are highlighted in \textbf{bold}, and the second best are \underline{underlined}.}
 \centering
 \begin{tabular}{lccccccccccc}
 \toprule
 \multirow{2}{*}{\centering Method} & \multicolumn{3}{c}{Synthetic Quality} & \multicolumn{5}{c}{LLVE} & \multicolumn{3}{c}{Object Detection} \\ 
 \cmidrule(lr){2-4} \cmidrule(lr){5-9} \cmidrule(lr){10-12}
 & KLD$\downarrow$ & FID$\downarrow$ & KID$\downarrow$ & PSNR$\uparrow$ & SSIM$\uparrow$ & LPIPS$\downarrow$ & CLIP-IQA$\uparrow$ & NIQE$\downarrow$ & AP$_{50-95}$(\%)$\uparrow$ & AP$_{50}$(\%)$\uparrow$ & AP$_{75}$(\%)$\uparrow$ \\
 \midrule
 AWGN & 0.458 & 258.635 & 0.450 & 29.716 & 0.849 & 0.165 &\underline{0.777} &6.234 & 18.474 & 20.328 & 19.812 \\
 P-G~\cite{foi2008noiseapprox} & \underline{0.358} & 254.760 & 0.444 & \underline{30.222} & 0.855 & \underline{0.134} & 0.763 & 5.943 & 25.213 & \underline{29.167} & 27.083 \\
 Starlight~\cite{monakhova2022starlight} & 0.444 & \underline{252.610} & \underline{0.428} & 30.006 & \underline{0.860} & 0.143 & \textbf{0.827} & \underline{5.812}& \underline{25.337} & 28.465 & \underline{28.465} \\
 Ours & \textbf{0.347} & \textbf{211.219} & \textbf{0.337} &\textbf{30.272} & \textbf{0.864} & \textbf{0.130} & 0.775 &\textbf{5.565} & \textbf{29.919} & \textbf{32.343} & \textbf{31.291} \\ 
 
 \bottomrule
 \end{tabular}
 \label{tab:results}
\end{table*}

\subsection{Synthetic Quality Assessment}
\label{ssec:synthetic_quality_assessment}

For a quantitative analysis of the performance of each synthetic pipeline, we calculate the Kullback–Leibler divergence (KLD), Fréchet Inception Distance (FID)~\cite{heusel2017fid} and Kernel Inception Distance (KID)~\cite{bińkowski2018kid} between the real noise maps and the synthetic noise maps. 
KL divergence is a broadly-applied statistical measure for quantifying the distance between probability distributions, and therefore is a highly suitable metric for our assessment.
FID and KID are two popular metrics used in image generation tasks for measuring the performance of generative models. FID measures the distance between two multivariate Gaussians fitted to feature representations of the input dataset against those of the target set. Unlike FID, KID does not assume a Gaussian distribution but instead uses Maximum Mean Discrepancy and is more suitable for small dataset sizes.

We use the SIDD-Small dataset~\cite{Abdelhamed2018SIDD} for evaluation as it covers a wide range of noise characteristics (which are influenced by various camera settings) across 5 different camera models. Moreover, the clean images are provided, allowing for the computation of real noise maps. For evaluation, we selected only the low-light scenes. The results, presented in Table \ref{tab:results} (under `Synthetic Quality'), demonstrate that our synthetic noise pipeline achieves the best results for all 3 metrics relative to the real noise, indicating a higher fidelity in noise replication. We would like to note that the FID scores are inflated, reflecting the metric’s strong bias to sample size.

\begin{figure}[!t]
 \small
 \centering
 \begin{minipage}[c]{0.04\linewidth}
  \centering 
  \rotatebox{90}{AWGN}
 \end{minipage}
 \begin{minipage}[c]{0.31\linewidth}
  \centering
  \includegraphics[width=\linewidth]{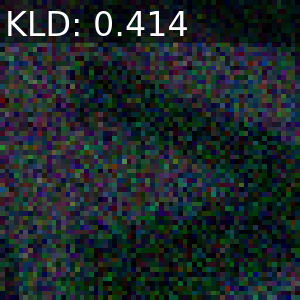}
 \end{minipage}
 \begin{minipage}[c]{0.31\linewidth}
  \centering
  \includegraphics[width=\linewidth]{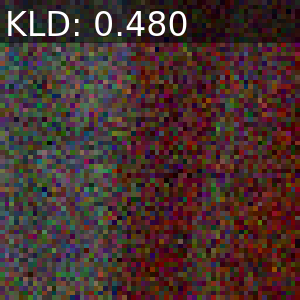}
 \end{minipage}
 \begin{minipage}[c]{0.31\linewidth}
  \centering
  \includegraphics[width=\linewidth]{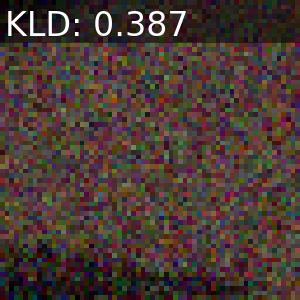}
 \end{minipage}
 \vspace{2pt}
 \\
 \noindent
 \begin{minipage}[c]{0.04\linewidth}
  \centering
  \rotatebox{90}{P-G~\cite{foi2008noiseapprox}}
 \end{minipage}
 \begin{minipage}[c]{0.31\linewidth}
  \centering
  \includegraphics[width=\linewidth]{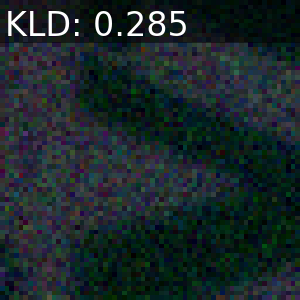}
 \end{minipage}
 \begin{minipage}[c]{0.31\linewidth}
  \centering
  \includegraphics[width=\linewidth]{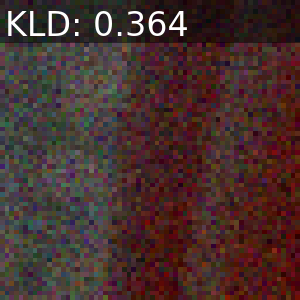}
 \end{minipage}
 \begin{minipage}[c]{0.31\linewidth}
  \centering
  \includegraphics[width=\linewidth]{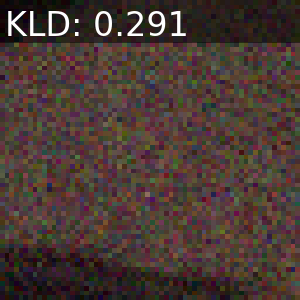}
 \end{minipage}
 \vspace{2pt}
 \\
 \noindent
 \begin{minipage}[c]{0.04\linewidth}
  \centering
  \rotatebox{90}{Starlight~\cite{monakhova2022starlight}}
 \end{minipage}
 \begin{minipage}[c]{0.31\linewidth}
  \centering
  \includegraphics[width=\linewidth]{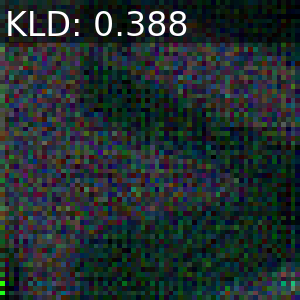}
 \end{minipage}
 \begin{minipage}[c]{0.31\linewidth}
  \centering
  \includegraphics[width=\linewidth]{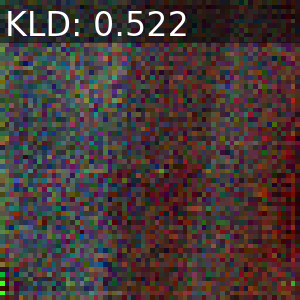}
 \end{minipage}
 \begin{minipage}[c]{0.31\linewidth}
  \centering
  \includegraphics[width=\linewidth]{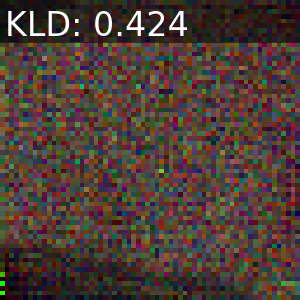}
 \end{minipage}
 \vspace{2pt}
 \\
 \noindent
 \begin{minipage}[c]{0.04\linewidth}
  \centering
  \rotatebox{90}{Ours}
 \end{minipage}
 \begin{minipage}[c]{0.31\linewidth}
  \centering
  \includegraphics[width=\linewidth]{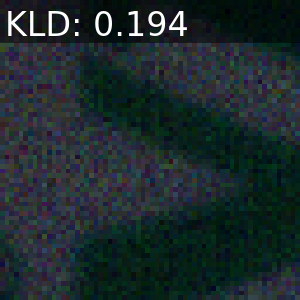}
 \end{minipage}
 \begin{minipage}[c]{0.31\linewidth}
  \centering
  \includegraphics[width=\linewidth]{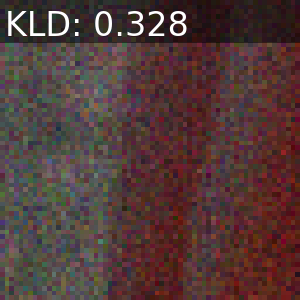}
 \end{minipage}
 \begin{minipage}[c]{0.31\linewidth}
  \centering
  \includegraphics[width=\linewidth]{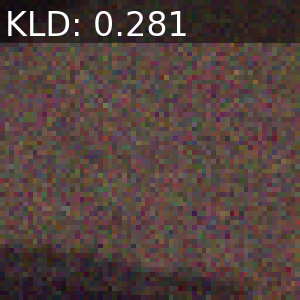}
 \end{minipage}
 \vspace{2pt}
 \\
 \noindent
 \begin{minipage}[c]{0.04\linewidth}
  \centering
  \rotatebox{90}{Real}
 \end{minipage}
 \begin{minipage}[c]{0.31\linewidth}
  \centering
  \includegraphics[width=\linewidth]{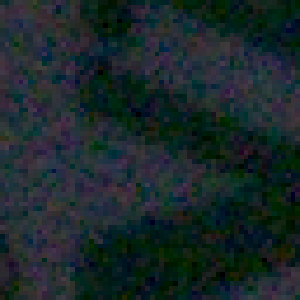}
 \end{minipage}
 \begin{minipage}[c]{0.31\linewidth}
  \centering
  \includegraphics[width=\linewidth]{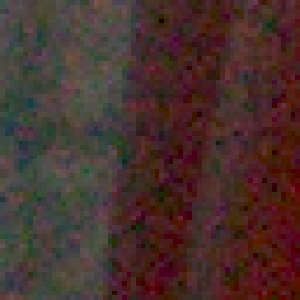}
 \end{minipage}
 \begin{minipage}[c]{0.31\linewidth}
  \centering
  \includegraphics[width=\linewidth]{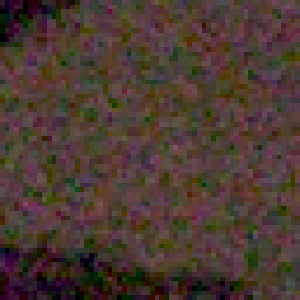}
 \end{minipage}
 \vspace{2pt}

 \caption{Visual comparison of different synthetic pipelines compared against the real low-light images from the SIDD-Small dataset~\cite{Abdelhamed2018SIDD}. The KL divergence for each synthetic image is also provided.}
 \Description{Examples of applying various synthetic noise pipelines onto patches from the SIDD-Small dataset, along with the KL divergence values, with ours visually appearing the most similar to the ground truth for every example.}
 \label{fig:visual_comparison_kld}
 \vspace{-1em}
\end{figure}

We also present visual comparisons of the results from each synthetic pipeline in Fig. \ref{fig:visual_comparison_kld} for qualitative analysis. Our method exhibits superior performance, adapting noise parameters dynamically to the noisy reference input. Subjectively, our results appear closer to the reference (noted as `Real' in the figure) and do not overestimate noise like other methods. However, as shown in Fig. \ref{fig:visual_comparison_kld}, there is evident noise reduction applied in the real noisy images which is not considered in our pipeline. These complex degradations will be addressed in our future work.

\subsection{Low-Light Video Enhancement (LLVE)}
\label{ssec:llve}

\begin{figure*}[t!]
 \centering
 \begin{minipage}[b]{0.5\linewidth}
  \centering
  \includegraphics[width=\linewidth]{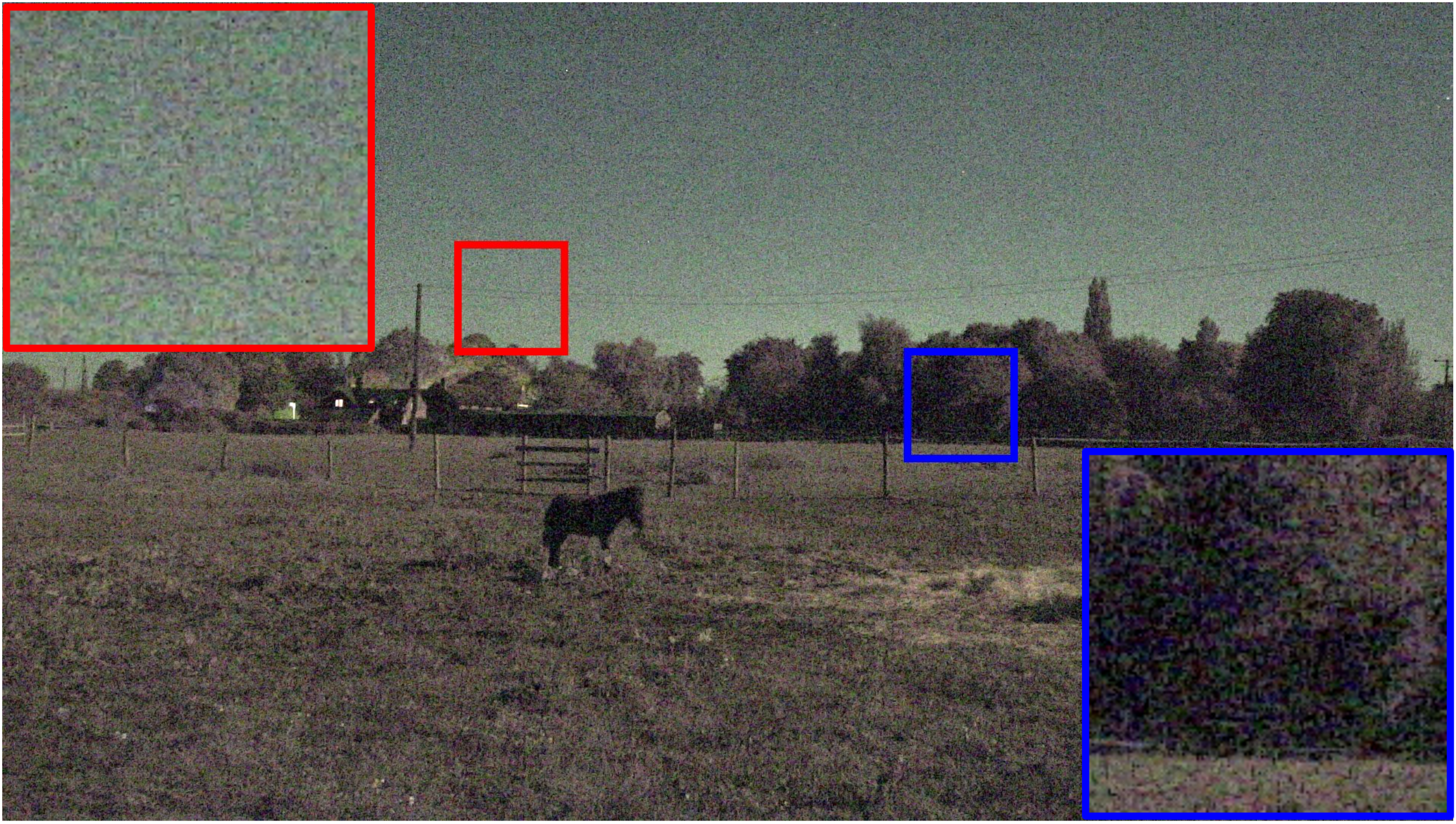}
 \end{minipage}
 \begin{minipage}[b]{0.49\linewidth}
  \begin{minipage}[c]{\linewidth}
   \includegraphics[width=0.24\linewidth]{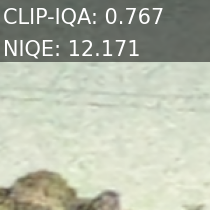}
   \includegraphics[width=0.24\linewidth]{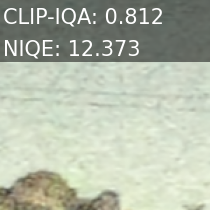}
   \includegraphics[width=0.24\linewidth]{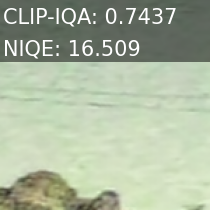}
   \includegraphics[width=0.24\linewidth]{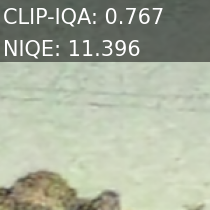}
  \end{minipage}%
  \vspace{3pt}
  \\
  \begin{minipage}[c]{\linewidth}
   \includegraphics[width=0.24\linewidth]{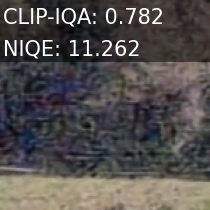}
   \includegraphics[width=0.24\linewidth]{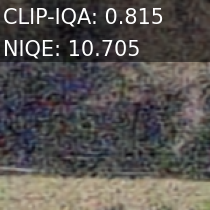}
   \includegraphics[width=0.24\linewidth]{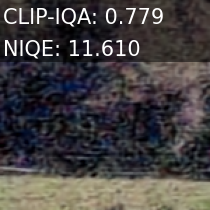}
   \includegraphics[width=0.24\linewidth]{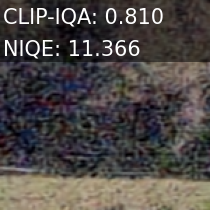}
  \end{minipage}%
  \vspace{5pt}
  \\
  \begin{minipage}[c][0.38cm][c]{\linewidth}
   \centering
   \begin{minipage}[c]{0.25\linewidth}
    \centering
    AWGN
   \end{minipage}%
   \begin{minipage}[c]{0.25\linewidth}
    \centering
    P-G~\cite{foi2008noiseapprox}
   \end{minipage}%
   \begin{minipage}[c]{0.25\linewidth}
    \centering
    Starlight~\cite{monakhova2022starlight}
   \end{minipage}%
   \begin{minipage}[c]{0.25\linewidth}
    \centering
    Ours
   \end{minipage}%
  \end{minipage}%
  \vspace{3pt}
 \end{minipage}%
 \caption{Qualitative comparison of the outputs of four BVI-Mamba~\cite{lin2024bvi} models trained with their respective noise pipelines on an extreme example of low-light noise. Simpler models are unable to capture the high degree of variability in the scene's noise, while more sophisticated models such as Starlight and ours allow BVI-Mamba to learn to denoise more effectively.}
 \Description{Example patches on a frame from the horse video, showing the denoising performances using BVI-Mamba trained on the various synthetic pipelines.}
 \label{fig:llve}
\end{figure*}



We assess the performance of our synthetic noise pipeline for low-light video enhancement through quantitative and qualitative analysis. For the low-light video enhancement experiments, we select a pre-trained model, BVI-Mamba~\cite{lin2024bvi}, as there are a limited number of video-specific LLVE methods. BVI-Mamba employs Visual State Space (VSS) blocks to reduce computational complexity. BVI-Mamba was pre-trained on the BVI-RLV~\cite{lin2024bvi} dataset, with over 30K fully-registered HD paired frames in normal/low-light conditions. We conduct our LLVE evaluation on a real noisy low-light video, which we refer to as the `horse' video. The `horse' video is captured outdoors after sunset with a Canon ML-105 camera (Fig.~\ref{fig:llve} left). For our synthetic pipeline, we first passed the horse video into DEN to estimate the noise parameters $\mathbf{v}_{horse}$, which is then used to generate the noisy videos for finetuning.

We first analyze how well the pre-trained low-light enhancer can learn different noise characteristics. We independently apply the same synthetic noise from each pipeline to the BVI-RLV testing sets and pass these synthesized inputs into the pre-trained BVI-Mamba model. The enhanced outputs are then compared with the clean, normal-light ground truths. Objective metrics include Peak Signal-to-Noise Ratio (PSNR), Structural Similarity Index Measure (SSIM), and perceptual similarity metrics like Learned Perceptual Image Patch Similarity (LPIPS)~\cite{zhang2018lpips}, are used to evaluate the performance of four different pipelines. The quantitative results can be seen in Table~\ref{tab:results}. 
The second analysis evaluates how closely the noise simulators can replicate the characteristics of the `horse' video. We train various BVI-Mamba models on different synthetic noise pipelines and evaluate their enhancing performances on the `horse' video. For our method, we first pass the `horse' video into DEN to estimate the noise parameters $\mathbf{v}_{horse}$, which are then used for synthesizing the noisy training data. As there is no clean ground-truth available for the `horse' video, we calculate no-reference metrics CLIP Image Quality Assessment (CLIP-IQA)~\cite{wang2022clipiqa} and Naturalness Image Quality Evaluator (NIQE)~\cite{mitall2013niqe}, for a quantitative comparison. Higher CLIP-IQA scores and lower NIQE scores indicate better perceptual quality. The prompts used for CLIP-IQA are `quality', `brightness', and `noisiness'. Visual comparisons are shown in Fig.~\ref{fig:llve}.

\begin{figure}[!t]

 \begin{minipage}[t]{0.32\linewidth}
  \centering
  \includegraphics[width=\linewidth]{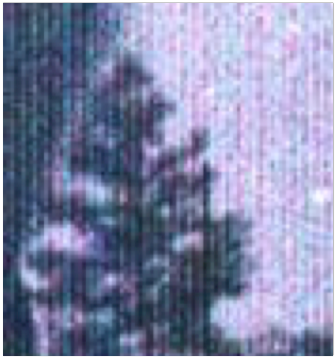}
 \end{minipage}%
 \hspace{1pt}
 \begin{minipage}[t]{0.32\linewidth}
  \centering
  \includegraphics[width=\linewidth]{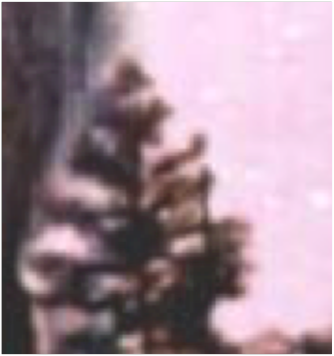}
 \end{minipage}%
  \hspace{1pt}
 \begin{minipage}[t]{0.32\linewidth}
  \centering
  \includegraphics[width=\linewidth]{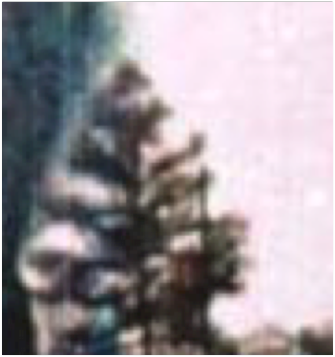}
 \end{minipage}%

 \begin{minipage}[c][0.38cm][c]{\linewidth}
  \centering
  \begin{minipage}[c]{0.32\linewidth}
   \centering
   Raw
  \end{minipage}%
  \begin{minipage}[c]{0.32\linewidth}
   \centering
   Starlight~\cite{monakhova2022starlight}
  \end{minipage}%
  \begin{minipage}[c]{0.32\linewidth}
   \centering
   Ours
  \end{minipage}%
 \end{minipage}

 \caption{Qualitative comparison of the outputs of BVI-Mamba models trained with receptive noise pipelines on an example patch from the Submillilux dataset~\cite{monakhova2022starlight}.}
 \Description{The denoising performances of the BVI-Mamba model trained on the different noise pipelinesn and tested on the Submillilux dataset.}
 \label{fig:submillilux_comparison}
 \vspace{-1.5em}
\end{figure}

Models trained with synthetic noise from our pipeline achieve a better balance between perceptual quality and naturalness. In smooth, bright regions, such as the sky shown in the top row, our method achieves the best NIQE scores compared to all other pipelines, while Starlight produces artifacts like color distortion. Although P-G has the highest CLIP-IQA score, our result visually exhibits less noise. In textured, dark areas, such as the tree region shown in the bottom row, P-G achieves better CLIP-IQA and NIQE scores, suggesting it might be better at enhancing brightness in those regions. However, results trained on our pipeline still maintain a competitive balance between detail restoration and denoising. We also compare enhancement performances on starlight videos from~\cite{monakhova2022starlight}, with BVI-Mamba trained on our synthetic pipeline shows competitive results~\ref{fig:submillilux_comparison}.

We further evaluate the enhancing capabilities of each synthetic pipeline by showing the improvements to object detection performances after enhancing the `horse' video. For this experiment, we use a pre-trained YOLOv11n model for detecting the horse in the video. Without any enhancement, the detection model was unable to detect the horse, which is expected as it was not trained on noisy data. After passing the `horse' video into the BVI-Mamba model trained with our synthetic pipeline, the pre-trained YOLOv11n model successfully detected the horse in the scene. In comparison, the BVI-Mamba model trained on the Starlight pipeline performed the worst by mislabeling the horse as a cow, while the model trained on AWGN noise resulted in two detections; a horse and a cow. Although the P-G data allows for successful horse detection, the confidence score is notably lower compared to its performance with our synthetic data.

\begin{figure}[!t]
 \centering

 \begin{minipage}[t]{0.49\linewidth}
  \centering
  \includegraphics[width=\linewidth]{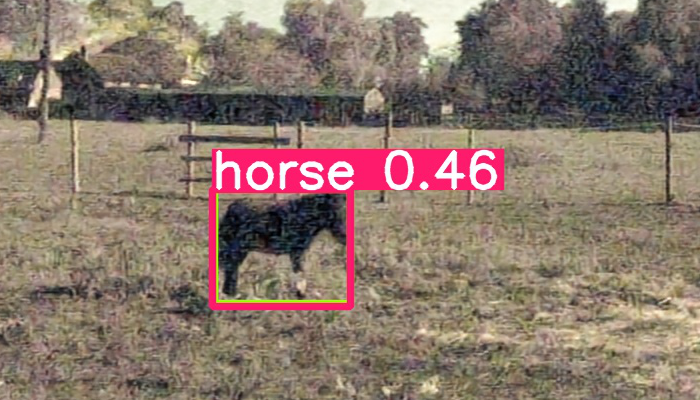}
 \end{minipage}%
 \hspace{1pt}
 \begin{minipage}[t]{0.49\linewidth}
  \centering
  \includegraphics[width=\linewidth]{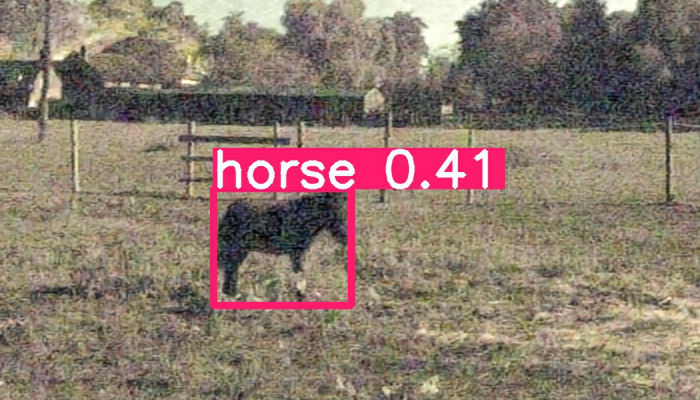}
 \end{minipage}%
 
 \begin{minipage}[c][0.38cm][c]{\linewidth}
  \centering
  \begin{minipage}[c]{0.49\linewidth}
   \centering
   AWGN
  \end{minipage}%
  \begin{minipage}[c]{0.49\linewidth}
   \centering
   P-G~\cite{foi2008noiseapprox}
  \end{minipage}%
 \end{minipage}
 \vspace{2pt}
 \begin{minipage}[t]{0.49\linewidth}
  \centering
  \includegraphics[width=\linewidth]{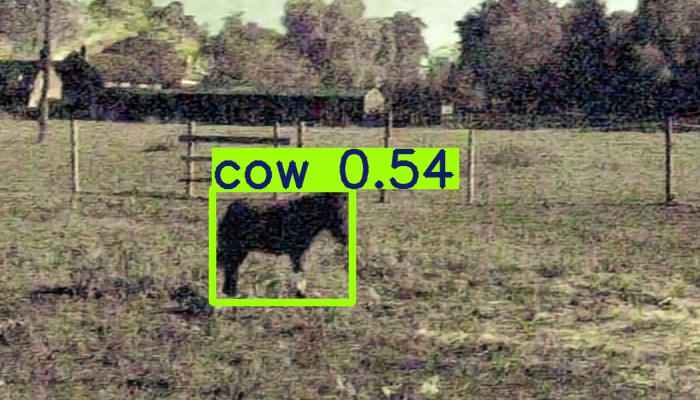}
 \end{minipage}%
 \hspace{1pt}
 \begin{minipage}[t]{0.49\linewidth}
  \centering
  \includegraphics[width=\linewidth]{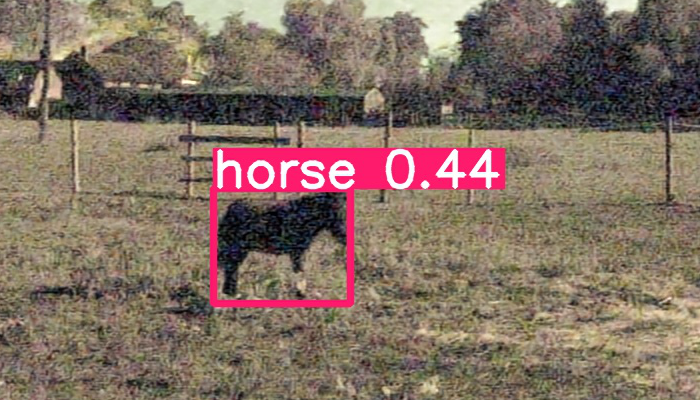}
 \end{minipage}%

 \begin{minipage}[c][0.38cm][c]{\linewidth}
  \centering
  \begin{minipage}[c]{0.49\linewidth}
   \centering
   Starlight~\cite{monakhova2022starlight}
  \end{minipage}%
  \begin{minipage}[c]{0.49\linewidth}
   \centering
   Ours
  \end{minipage}%
 \end{minipage}

 \caption{Qualitative comparison of the pre-trained YOLOv11n~\cite{Jocher2023yolo} on outputs of four BVI-Mamba models trained with their respective noise pipelines. The denoiser trained using our approach has the most successful result.}
 \Description{The detection performances of a pre-trained YOLOv11n model on the denoised outputs from the BVI-Mamba models trained on the various noise pipelines.}
 \label{fig:horse_detection}
 \vspace{-2em}
\end{figure}
\subsection{Object Detection}

\begin{figure*}[!t]
    \begin{center}
        \includegraphics[width=0.49\linewidth]{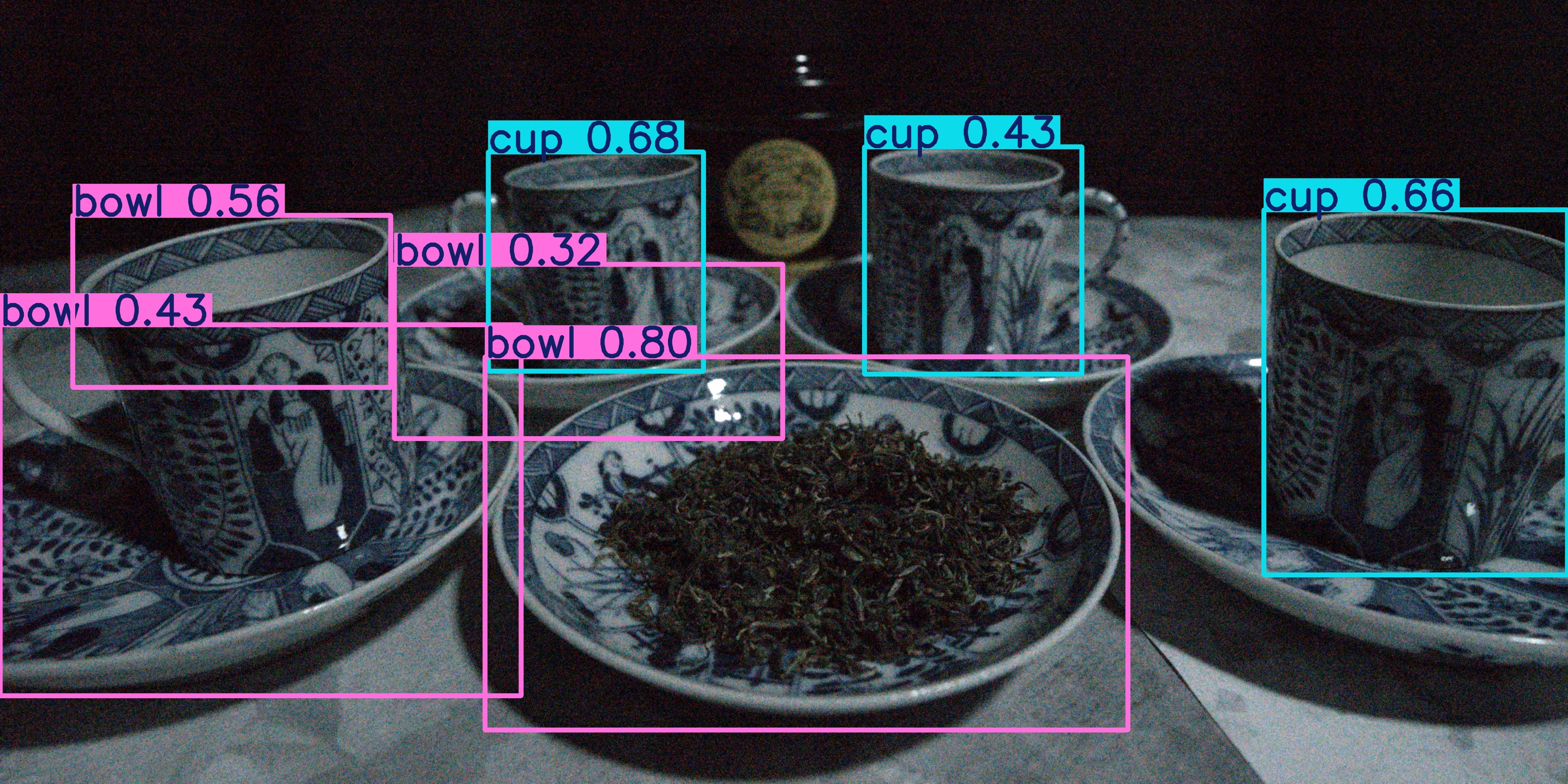}\hfill
        \includegraphics[width=0.49\linewidth]{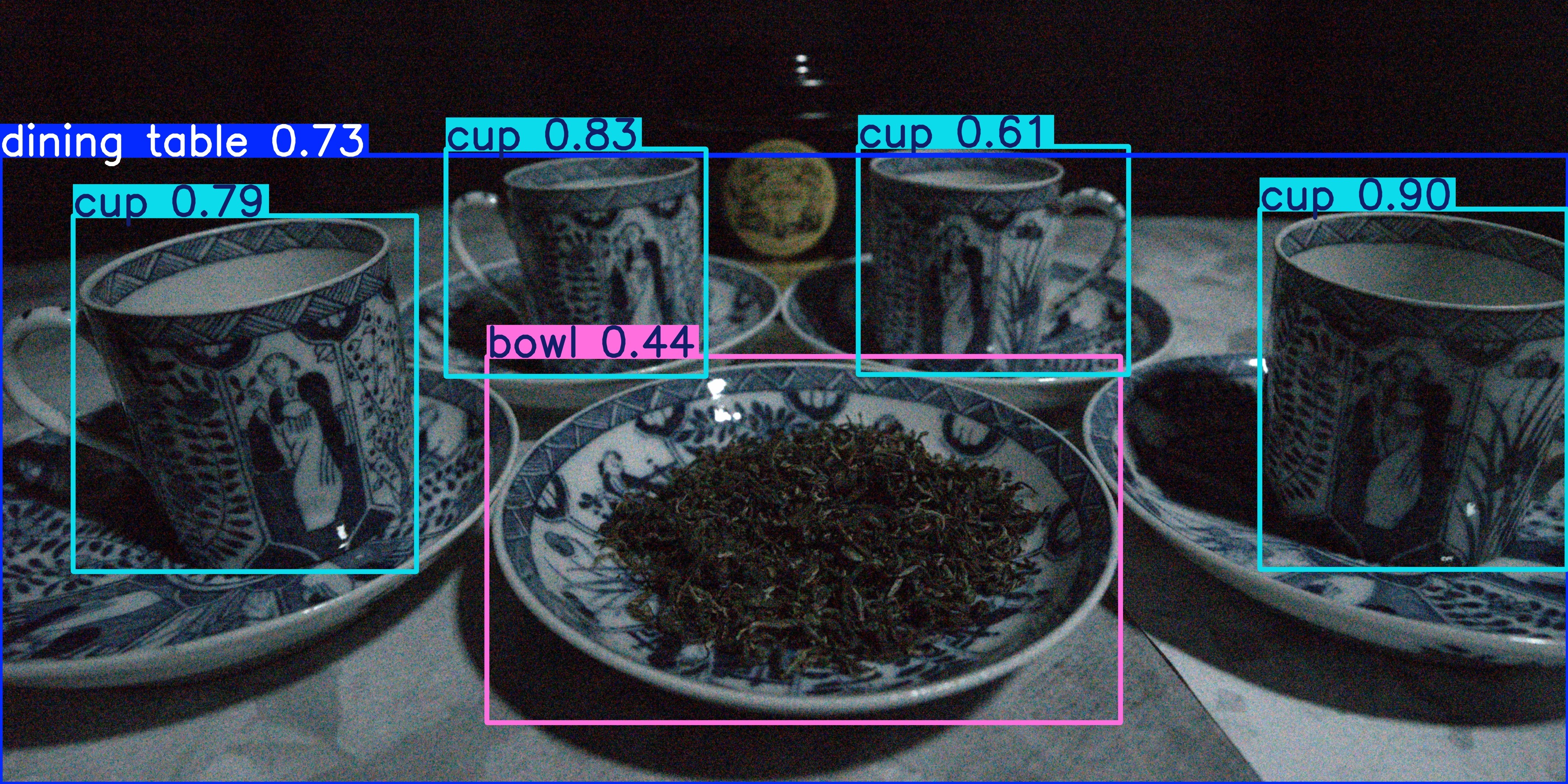} \hfill
    \end{center}

 \caption{Qualitative comparison showing the performance improvements when using our synthetic noise pipeline to train an object detection model for use in low-light conditions on an image from BVI-Lowlight dataset~\cite{anantrasirichai2024bvi}. Left: pre-trained YOLOv11n~\cite{Jocher2023yolo}. Right: YOLOv11n trained with our synthetic data.}
 \Description{Comparison photos of object detection performances using a pre-trained YOLOv11n model and a YOLOv11n model trained on our synthetic data.}
 \label{fig:object_detection}
\end{figure*}

To determine the effectiveness of our synthetic pipeline for downstream tasks separate to video enhancement, we conduct object detection experiments under real low-light conditions. For each of the four different synthetic pipelines, we train a YOLOv11n~\cite{Jocher2023yolo} model on the popular Microsoft COCO dataset~\cite{lin2014coco}. The synthetic noise is inserted \textit{after} simulating low-light conditions, using Equation~\ref{eq:lowlight}. Each model is trained on its respective synthetic low-light data for 100 epochs.

The models are evaluated on the highest ISO frames from the real low-light image dataset~\cite{anantrasirichai2024bvi}, which exhibits higher noise levels relative to the other existing real low-light datasets. The performance of each model is tested using pseudo-ground-truth annotations generated by applying the pre-trained YOLOv11n onto the clean ISO-100 videos. In 4 of the 20 scenes, the pre-trained model did not detect any instances of the 80 COCO classes, so those scenes are removed from the test set. We measured this performance using the standard object detection metric Average Precision (AP), over various Intersection over Union (IoU) thresholds; AP$_{50-95}$, AP$_{50}$, and AP$_{75}$.

As shown in Table~\ref{tab:results} (under `Object Detection'), training the YOLOv11n model using our training approach results in improved precision across all thresholds. This demonstrates that our synthetic noise is capable of reproducing the noise characteristics of the dataset, and suitable for low-light domain adaptation. Examples of detection results are shown in Fig.~\ref{fig:object_detection}; without training directly on noisy images, the pre-trained YOLOv11n struggled to differentiate cups and bowls, and ignores the presence of a dining table. However, training on our synthetic pipeline, resulted in much greater performance, with correct classification of the cups and less false positive detections.

It is noted that the pre-trained YOLOv11n applied to the clean image (for pseudo ground-truth) is not without fault. It failed to detect the other four bowls in the image with high confidence, and only exhibited particularly high confidence for the cups in the scene. In addition, these three cases exhibited higher performance due to the relatively high light conditions.

\subsection{Ablation Study}

We assess the significance of each component in our network architecture through an ablation study, showing the results in Table~\ref{tab:results_ablation}. We conduct 3 experiments, each with a different combination of the components, calculating KLD, FID and KID of the noise maps for each experiment.
We refer to the different architectures as model \textit{v1}, \textit{v2}, and \textit{v3}.
Model \textit{v1} is simply a U-Net~\cite{ronneberge2015unet} trained to generate a synthetic noisy image without employing $f(\cdot)$.
Model \textit{v2} resembles a classification network, with an encoder and a MLP head for estimating the noise parameters.
Model \textit{v3}, the proposed architecture for DEN, is comprised of an encoder, a decoder and an MLP head, allowing for the network to understand the noise characteristics further via the dual network heads.

We highlight the importance of our MLP head, demonstrating a major performance increase when added to the U-Net, with a substantial improvement of 382.916 in FID score between model \textit{v1} and model \textit{v2}, and therefore it is a crucial component for allowing our network to mimic the noise characteristics accurately.
Furthermore, we verify that our $\mathcal{L}_{rec}$ improves our network performance by assisting in understanding noise in an image/video; the added decoder head and accompanying $\mathcal{L}_{rec}$, forces the model to learn the specifics of the reference noise, as seen in the increase across all metrics for model \textit{v3}.
This ablation study confirms the significance of each component in our proposed DEN.

\begin{table}[!t]
 \centering
 \caption{Ablation Study of the DEN Architecture.}
 \begin{tabular}{ccccccc}
 \toprule
 Model & Encoder & Decoder & MLP & KLD$\downarrow$ & FID$\downarrow$ & KID$\downarrow$\\
 \midrule
 \textit{v1} & \cmark & \cmark & \xmark & 0.425 & 604.363 & 1.043 \\
 \textit{v2} & \cmark & \xmark & \cmark & 0.379 & 221.447 & 0.361	 \\ 
 \textit{v3} & \cmark & \cmark & \cmark & \textbf{0.347} & \textbf{211.219} & \textbf{0.337} \\ 
 \bottomrule
 \end{tabular}
 \label{tab:results_ablation}
\end{table}


\section{Conclusion}

We propose a novel general-purpose zero-shot synthetic low-light image and video pipeline which is capable of synthesizing realistic noise from unseen real low-light sources. Unlike other synthetic sRGB deep-learning pipelines, our Degradation Estimation Network does not need information regarding the video capture, enabling seamless application in downstream tasks. Furthermore, our method is applicable to a diverse range of noise characteristics without the need for training targeting individual datasets; making our method easier to generate realistic low-light videos for training deep-learning models. We verify the robustness of our pipeline with our comprehensive study, achieving improvements up to 24\% KLD, 21\% LPIPS, and 62\% AP$_{50-95}$.

In our future work, we will explore video-specific degradations, such as motion blur, as most existing synthetic pipelines focus on image-only degradations. We would like to also address more complex degradations, like spatially-correlated noise, arising from the camera ISP. Furthermore, our model currently focuses on noise synthesis in low-light videos; we would like to expand our method to learn the illumination mapping and color distortions as well, allowing easier low-light synthesis for downstream tasks. Finally, we aim to explore using our method for joint low-light enhancement and synthesis by improving DEN's reconstruction capabilities.


\begin{acks}
This work was supported by the UKRI MyWorld Strength in Places Programme (SIPF00006/1), the EPSRC Doctoral Training Partnerships (EP/W524414/1), and an EPSRC iCASE grant in collaboration with the British Broadcasting Corporation (BBC).
\end{acks}

\bibliographystyle{ACM-Reference-Format}
\bibliography{refs}


\begin{thebibliography}{45}


\ifx \showCODEN    \undefined \def \showCODEN     #1{\unskip}     \fi
\ifx \showISBNx    \undefined \def \showISBNx     #1{\unskip}     \fi
\ifx \showISBNxiii \undefined \def \showISBNxiii  #1{\unskip}     \fi
\ifx \showISSN     \undefined \def \showISSN      #1{\unskip}     \fi
\ifx \showLCCN     \undefined \def \showLCCN      #1{\unskip}     \fi
\ifx \shownote     \undefined \def \shownote      #1{#1}          \fi
\ifx \showarticletitle \undefined \def \showarticletitle #1{#1}   \fi
\ifx \showURL      \undefined \def \showURL       {\relax}        \fi
\providecommand\bibfield[2]{#2}
\providecommand\bibinfo[2]{#2}
\providecommand\natexlab[1]{#1}
\providecommand\showeprint[2][]{arXiv:#2}

\bibitem[Abdelhamed et~al\mbox{.}(2018)]%
        {Abdelhamed2018SIDD}
\bibfield{author}{\bibinfo{person}{Abdelrahman Abdelhamed}, \bibinfo{person}{Stephen Lin}, {and} \bibinfo{person}{Michael~S. Brown}.} \bibinfo{year}{2018}\natexlab{}.
\newblock \showarticletitle{A High-Quality Denoising Dataset for Smartphone Cameras}. In \bibinfo{booktitle}{\emph{IEEE Conference on Computer Vision and Pattern Recognition (CVPR)}}.
\newblock


\bibitem[Anantrasirichai et~al\mbox{.}(2015)]%
        {anantrasirichai2015cwttexture}
\bibfield{author}{\bibinfo{person}{Nantheera. Anantrasirichai}, \bibinfo{person}{Jeremy. Burn}, {and} \bibinfo{person}{David~R. Bull}.} \bibinfo{year}{2015}\natexlab{}.
\newblock \showarticletitle{Robust texture features based on undecimated dual-tree complex wavelets and local magnitude binary patterns}. In \bibinfo{booktitle}{\emph{2015 IEEE International Conference on Image Processing (ICIP)}}. \bibinfo{pages}{3957--3961}.
\newblock
\href{https://doi.org/10.1109/ICIP.2015.7351548}{doi:\nolinkurl{10.1109/ICIP.2015.7351548}}


\bibitem[Anantrasirichai et~al\mbox{.}(2024)]%
        {anantrasirichai2024bvi}
\bibfield{author}{\bibinfo{person}{Nantheera Anantrasirichai}, \bibinfo{person}{Ruirui Lin}, \bibinfo{person}{Alexandra Malyugina}, {and} \bibinfo{person}{David Bull}.} \bibinfo{year}{2024}\natexlab{}.
\newblock \showarticletitle{BVI-Lowlight: Fully registered benchmark dataset for low-light video enhancement}.
\newblock \bibinfo{journal}{\emph{arXiv preprint arXiv:2402.01970}} (\bibinfo{year}{2024}).
\newblock


\bibitem[Anantrasirichai et~al\mbox{.}(2025)]%
        {anantrasirichai2025AI}
\bibfield{author}{\bibinfo{person}{Nantheera Anantrasirichai}, \bibinfo{person}{Fan Zhang}, {and} \bibinfo{person}{David Bull}.} \bibinfo{year}{2025}\natexlab{}.
\newblock \bibinfo{title}{Artificial Intelligence in Creative Industries: Advances Prior to 2025}.
\newblock
\showeprint[arxiv]{2501.02725}~[cs.AI]
\urldef\tempurl%
\url{https://arxiv.org/abs/2501.02725}
\showURL{%
\tempurl}


\bibitem[Bińkowski et~al\mbox{.}(2018)]%
        {bińkowski2018kid}
\bibfield{author}{\bibinfo{person}{Mikołaj Bińkowski}, \bibinfo{person}{Dougal~J. Sutherland}, \bibinfo{person}{Michael Arbel}, {and} \bibinfo{person}{Arthur Gretton}.} \bibinfo{year}{2018}\natexlab{}.
\newblock \showarticletitle{Demystifying {MMD} {GAN}s}. In \bibinfo{booktitle}{\emph{International Conference on Learning Representations}}.
\newblock
\urldef\tempurl%
\url{https://openreview.net/forum?id=r1lUOzWCW}
\showURL{%
\tempurl}


\bibitem[Boncelet(2009)]%
        {boncelet2009noisemodels}
\bibfield{author}{\bibinfo{person}{Charles Boncelet}.} \bibinfo{year}{2009}\natexlab{}.
\newblock \showarticletitle{Chapter 7 - Image Noise Models}.
\newblock In \bibinfo{booktitle}{\emph{The Essential Guide to Image Processing}}, \bibfield{editor}{\bibinfo{person}{Al~Bovik}} (Ed.). \bibinfo{publisher}{Academic Press}, \bibinfo{address}{Boston}, \bibinfo{pages}{143--167}.
\newblock
\showISBNx{978-0-12-374457-9}
\href{https://doi.org/10.1016/B978-0-12-374457-9.00007-X}{doi:\nolinkurl{10.1016/B978-0-12-374457-9.00007-X}}


\bibitem[Cao et~al\mbox{.}(2023)]%
        {Cao2023ISOdependentnoise}
\bibfield{author}{\bibinfo{person}{Yue Cao}, \bibinfo{person}{Ming Liu}, \bibinfo{person}{Shuai Liu}, \bibinfo{person}{Xiaotao Wang}, \bibinfo{person}{Lei Lei}, {and} \bibinfo{person}{Wangmeng Zuo}.} \bibinfo{year}{2023}\natexlab{}.
\newblock \showarticletitle{Physics-Guided ISO-Dependent Sensor Noise Modeling for Extreme Low-Light Photography}. In \bibinfo{booktitle}{\emph{Proceedings of the IEEE/CVF Conference on Computer Vision and Pattern Recognition (CVPR)}}. \bibinfo{pages}{5744--5753}.
\newblock


\bibitem[Chen and Ma(2025)]%
        {chen2025qualityandperception}
\bibfield{author}{\bibinfo{person}{Hongyang Chen} {and} \bibinfo{person}{Kaisheng Ma}.} \bibinfo{year}{2025}\natexlab{}.
\newblock \showarticletitle{Enhancing Vision: Harmonizing Frequency for Imaging Quality and Perception Accuracy}. In \bibinfo{booktitle}{\emph{ICASSP 2025 - 2025 IEEE International Conference on Acoustics, Speech and Signal Processing (ICASSP)}}. \bibinfo{pages}{1--5}.
\newblock
\showISSN{2379-190X}
\href{https://doi.org/10.1109/ICASSP49660.2025.10889903}{doi:\nolinkurl{10.1109/ICASSP49660.2025.10889903}}


\bibitem[Cui et~al\mbox{.}(2021)]%
        {cui2021maet}
\bibfield{author}{\bibinfo{person}{Ziteng Cui}, \bibinfo{person}{Guo-Jun Qi}, \bibinfo{person}{Lin Gu}, \bibinfo{person}{Shaodi You}, \bibinfo{person}{Zenghui Zhang}, {and} \bibinfo{person}{Tatsuya Harada}.} \bibinfo{year}{2021}\natexlab{}.
\newblock \showarticletitle{Multitask AET With Orthogonal Tangent Regularity for Dark Object Detection}. In \bibinfo{booktitle}{\emph{Proceedings of the IEEE/CVF International Conference on Computer Vision (ICCV)}}. \bibinfo{pages}{2553--2562}.
\newblock


\bibitem[Damen et~al\mbox{.}(2018)]%
        {Damen2018EPICKITCHENS}
\bibfield{author}{\bibinfo{person}{Dima Damen}, \bibinfo{person}{Hazel Doughty}, \bibinfo{person}{Giovanni~Maria Farinella}, \bibinfo{person}{Sanja Fidler}, \bibinfo{person}{Antonino Furnari}, \bibinfo{person}{Evangelos Kazakos}, \bibinfo{person}{Davide Moltisanti}, \bibinfo{person}{Jonathan Munro}, \bibinfo{person}{Toby Perrett}, \bibinfo{person}{Will Price}, {and} \bibinfo{person}{Michael Wray}.} \bibinfo{year}{2018}\natexlab{}.
\newblock \showarticletitle{Scaling Egocentric Vision: The EPIC-KITCHENS Dataset}. In \bibinfo{booktitle}{\emph{European Conference on Computer Vision (ECCV)}}.
\newblock


\bibitem[El~Gamal and Eltoukhy(2005)]%
        {ElGamal2005cmos}
\bibfield{author}{\bibinfo{person}{A. El~Gamal} {and} \bibinfo{person}{H. Eltoukhy}.} \bibinfo{year}{2005}\natexlab{}.
\newblock \showarticletitle{CMOS image sensors}.
\newblock \bibinfo{journal}{\emph{IEEE Circuits and Devices Magazine}} \bibinfo{volume}{21}, \bibinfo{number}{3} (\bibinfo{year}{2005}), \bibinfo{pages}{6--20}.
\newblock
\href{https://doi.org/10.1109/MCD.2005.1438751}{doi:\nolinkurl{10.1109/MCD.2005.1438751}}


\bibitem[Foi et~al\mbox{.}(2008)]%
        {foi2008noiseapprox}
\bibfield{author}{\bibinfo{person}{Alessandro Foi}, \bibinfo{person}{Mejdi Trimeche}, \bibinfo{person}{Vladimir Katkovnik}, {and} \bibinfo{person}{Karen Egiazarian}.} \bibinfo{year}{2008}\natexlab{}.
\newblock \showarticletitle{Practical Poissonian-Gaussian Noise Modeling and Fitting for Single-Image Raw-Data}.
\newblock \bibinfo{journal}{\emph{IEEE Transactions on Image Processing}} \bibinfo{volume}{17}, \bibinfo{number}{10} (\bibinfo{year}{2008}), \bibinfo{pages}{1737--1754}.
\newblock
\href{https://doi.org/10.1109/TIP.2008.2001399}{doi:\nolinkurl{10.1109/TIP.2008.2001399}}


\bibitem[Fu et~al\mbox{.}(2023b)]%
        {fu2023did}
\bibfield{author}{\bibinfo{person}{Huiyuan Fu}, \bibinfo{person}{Wenkai Zheng}, \bibinfo{person}{Xicong Wang}, \bibinfo{person}{Jiaxuan Wang}, \bibinfo{person}{Heng Zhang}, {and} \bibinfo{person}{Huadong Ma}.} \bibinfo{year}{2023}\natexlab{b}.
\newblock \showarticletitle{Dancing in the Dark: A Benchmark towards General Low-light Video Enhancement}. In \bibinfo{booktitle}{\emph{Proceedings of the IEEE/CVF International Conference on Computer Vision (ICCV)}}. \bibinfo{pages}{12877--12886}.
\newblock


\bibitem[Fu et~al\mbox{.}(2023a)]%
        {fu2023sRGBneighbornoise}
\bibfield{author}{\bibinfo{person}{Zixuan Fu}, \bibinfo{person}{Lanqing Guo}, {and} \bibinfo{person}{Bihan Wen}.} \bibinfo{year}{2023}\natexlab{a}.
\newblock \showarticletitle{sRGB Real Noise Synthesizing With Neighboring Correlation-Aware Noise Model}. In \bibinfo{booktitle}{\emph{Proceedings of the IEEE/CVF Conference on Computer Vision and Pattern Recognition (CVPR)}}. \bibinfo{pages}{1683--1691}.
\newblock


\bibitem[Guo et~al\mbox{.}(2020)]%
        {guo2020zerodce}
\bibfield{author}{\bibinfo{person}{Chunle~Guo Guo}, \bibinfo{person}{Chongyi Li}, \bibinfo{person}{Jichang Guo}, \bibinfo{person}{Chen~Change Loy}, \bibinfo{person}{Junhui Hou}, \bibinfo{person}{Sam Kwong}, {and} \bibinfo{person}{Runmin Cong}.} \bibinfo{year}{2020}\natexlab{}.
\newblock \showarticletitle{Zero-reference deep curve estimation for low-light image enhancement}. In \bibinfo{booktitle}{\emph{Proceedings of the IEEE conference on computer vision and pattern recognition (CVPR)}}. \bibinfo{pages}{1780--1789}.
\newblock


\bibitem[Heusel et~al\mbox{.}(2017)]%
        {heusel2017fid}
\bibfield{author}{\bibinfo{person}{Martin Heusel}, \bibinfo{person}{Hubert Ramsauer}, \bibinfo{person}{Thomas Unterthiner}, \bibinfo{person}{Bernhard Nessler}, {and} \bibinfo{person}{Sepp Hochreiter}.} \bibinfo{year}{2017}\natexlab{}.
\newblock \showarticletitle{GANs Trained by a Two Time-Scale Update Rule Converge to a Local Nash Equilibrium}. In \bibinfo{booktitle}{\emph{Advances in Neural Information Processing Systems}}, \bibfield{editor}{\bibinfo{person}{I.~Guyon}, \bibinfo{person}{U.~Von Luxburg}, \bibinfo{person}{S.~Bengio}, \bibinfo{person}{H.~Wallach}, \bibinfo{person}{R.~Fergus}, \bibinfo{person}{S.~Vishwanathan}, {and} \bibinfo{person}{R.~Garnett}} (Eds.), Vol.~\bibinfo{volume}{30}. \bibinfo{publisher}{Curran Associates, Inc.}
\newblock
\urldef\tempurl%
\url{https://proceedings.neurips.cc/paper_files/paper/2017/file/8a1d694707eb0fefe65871369074926d-Paper.pdf}
\showURL{%
\tempurl}


\bibitem[Huang et~al\mbox{.}(2021)]%
        {huang2021got10k}
\bibfield{author}{\bibinfo{person}{Lianghua Huang}, \bibinfo{person}{Xin Zhao}, {and} \bibinfo{person}{Kaiqi Huang}.} \bibinfo{year}{2021}\natexlab{}.
\newblock \showarticletitle{GOT-10k: A Large High-Diversity Benchmark for Generic Object Tracking in the Wild}.
\newblock \bibinfo{journal}{\emph{IEEE Transactions on Pattern Analysis and Machine Intelligence}} \bibinfo{volume}{43}, \bibinfo{number}{5} (\bibinfo{year}{2021}), \bibinfo{pages}{1562--1577}.
\newblock
\href{https://doi.org/10.1109/TPAMI.2019.2957464}{doi:\nolinkurl{10.1109/TPAMI.2019.2957464}}


\bibitem[Jocher et~al\mbox{.}(2023)]%
        {Jocher2023yolo}
\bibfield{author}{\bibinfo{person}{Glenn Jocher}, \bibinfo{person}{Jing Qiu}, {and} \bibinfo{person}{Ayush Chaurasia}.} \bibinfo{year}{2023}\natexlab{}.
\newblock \bibinfo{title}{{Ultralytics YOLO}}.
\newblock
\urldef\tempurl%
\url{https://github.com/ultralytics/ultralytics}
\showURL{%
\tempurl}


\bibitem[Kousha et~al\mbox{.}(2022)]%
        {kousha2022sRGBnormflow}
\bibfield{author}{\bibinfo{person}{Shayan Kousha}, \bibinfo{person}{Ali Maleky}, \bibinfo{person}{Michael~S. Brown}, {and} \bibinfo{person}{Marcus~A. Brubaker}.} \bibinfo{year}{2022}\natexlab{}.
\newblock \showarticletitle{Modeling sRGB Camera Noise With Normalizing Flows}. In \bibinfo{booktitle}{\emph{Proceedings of the IEEE/CVF Conference on Computer Vision and Pattern Recognition (CVPR)}}. \bibinfo{pages}{17463--17471}.
\newblock


\bibitem[Li et~al\mbox{.}(2024)]%
        {Li2024LLEVOS}
\bibfield{author}{\bibinfo{person}{Hebei Li}, \bibinfo{person}{Jin Wang}, \bibinfo{person}{Jiahui Yuan}, \bibinfo{person}{Yue Li}, \bibinfo{person}{Wenming Weng}, \bibinfo{person}{Yansong Peng}, \bibinfo{person}{Yueyi Zhang}, \bibinfo{person}{Zhiwei Xiong}, {and} \bibinfo{person}{Xiaoyan Sun}.} \bibinfo{year}{2024}\natexlab{}.
\newblock \showarticletitle{Event-assisted Low-Light Video Object Segmentation}. In \bibinfo{booktitle}{\emph{Proceedings of the IEEE/CVF Conference on Computer Vision and Pattern Recognition (CVPR)}}. \bibinfo{pages}{3250--3259}.
\newblock


\bibitem[Lin et~al\mbox{.}(2025)]%
        {lin2025segmlowlight}
\bibfield{author}{\bibinfo{person}{Joanne Lin}, \bibinfo{person}{Nantheera Anantrasirichai}, {and} \bibinfo{person}{David Bull}.} \bibinfo{year}{2025}\natexlab{}.
\newblock \showarticletitle{Multi-Scale Denoising in the Feature Space for Low-Light Instance Segmentation}. In \bibinfo{booktitle}{\emph{ICASSP 2025 - 2025 IEEE International Conference on Acoustics, Speech and Signal Processing (ICASSP)}}. \bibinfo{pages}{1--5}.
\newblock
\href{https://doi.org/10.1109/ICASSP49660.2025.10889336}{doi:\nolinkurl{10.1109/ICASSP49660.2025.10889336}}


\bibitem[Lin et~al\mbox{.}(2024a)]%
        {lin2024bvi}
\bibfield{author}{\bibinfo{person}{Ruiui Lin}, \bibinfo{person}{Nantheera Anantrasirichai}, \bibinfo{person}{Guoxi Huang}, \bibinfo{person}{Joanne Lin}, \bibinfo{person}{Qi Sun}, \bibinfo{person}{Alexandra Malyugina}, {and} \bibinfo{person}{David~R Bull}.} \bibinfo{year}{2024}\natexlab{a}.
\newblock \showarticletitle{{BVI-RLV}: A Fully Registered Dataset and Benchmarks for Low-Light Video Enhancement}.
\newblock \bibinfo{journal}{\emph{arXiv preprint arXiv:2407.03535}} (\bibinfo{year}{2024}).
\newblock


\bibitem[Lin et~al\mbox{.}(2024b)]%
        {lin2024stasunet}
\bibfield{author}{\bibinfo{person}{Ruirui Lin}, \bibinfo{person}{Nantheera Anantrasirichai}, \bibinfo{person}{Alexandra Malyugina}, {and} \bibinfo{person}{David Bull}.} \bibinfo{year}{2024}\natexlab{b}.
\newblock \showarticletitle{A Spatio-Temporal Aligned SUNet Model For Low-Light Video Enhancement}. In \bibinfo{booktitle}{\emph{2024 IEEE International Conference on Image Processing (ICIP)}}. \bibinfo{pages}{1480--1486}.
\newblock
\href{https://doi.org/10.1109/ICIP51287.2024.10647380}{doi:\nolinkurl{10.1109/ICIP51287.2024.10647380}}


\bibitem[Lin et~al\mbox{.}(2014)]%
        {lin2014coco}
\bibfield{author}{\bibinfo{person}{Tsung-Yi Lin}, \bibinfo{person}{Michael Maire}, \bibinfo{person}{Serge Belongie}, \bibinfo{person}{James Hays}, \bibinfo{person}{Pietro Perona}, \bibinfo{person}{Deva Ramanan}, \bibinfo{person}{Piotr Doll{\'a}r}, {and} \bibinfo{person}{C~Lawrence Zitnick}.} \bibinfo{year}{2014}\natexlab{}.
\newblock \showarticletitle{Microsoft COCO: Common Objects in Context}. In \bibinfo{booktitle}{\emph{European Conference on Computer Vision}}. \bibinfo{pages}{740--755}.
\newblock


\bibitem[Liu et~al\mbox{.}(2024)]%
        {liu2024ntvot211}
\bibfield{author}{\bibinfo{person}{Yu Liu}, \bibinfo{person}{Arif Mahmood}, {and} \bibinfo{person}{Muhammad~Haris Khan}.} \bibinfo{year}{2024}\natexlab{}.
\newblock \showarticletitle{NT-VOT211: A Large-Scale Benchmark for Night-time Visual Object Tracking}. In \bibinfo{booktitle}{\emph{Proceedings of the Asian Conference on Computer Vision (ACCV)}}. \bibinfo{pages}{194--212}.
\newblock


\bibitem[Luo et~al\mbox{.}(2023)]%
        {luo2023similarityminmax}
\bibfield{author}{\bibinfo{person}{Rundong Luo}, \bibinfo{person}{Wenjing Wang}, \bibinfo{person}{Wenhan Yang}, {and} \bibinfo{person}{Jiaying Liu}.} \bibinfo{year}{2023}\natexlab{}.
\newblock \showarticletitle{Similarity Min-Max: Zero-Shot Day-Night Domain Adaptation}. In \bibinfo{booktitle}{\emph{ICCV}}.
\newblock


\bibitem[Lv et~al\mbox{.}(2021)]%
        {lv2021agllnet}
\bibfield{author}{\bibinfo{person}{Feifan Lv}, \bibinfo{person}{Yu Li}, {and} \bibinfo{person}{Feng Lu}.} \bibinfo{year}{2021}\natexlab{}.
\newblock \showarticletitle{Attention guided low-light image enhancement with a large scale low-light simulation dataset}.
\newblock \bibinfo{journal}{\emph{International Journal of Computer Vision}} \bibinfo{volume}{129}, \bibinfo{number}{7} (\bibinfo{year}{2021}), \bibinfo{pages}{2175--2193}.
\newblock


\bibitem[Mittal et~al\mbox{.}(2013)]%
        {mitall2013niqe}
\bibfield{author}{\bibinfo{person}{Anish Mittal}, \bibinfo{person}{Rajiv Soundararajan}, {and} \bibinfo{person}{Alan~C. Bovik}.} \bibinfo{year}{2013}\natexlab{}.
\newblock \showarticletitle{Making a “Completely Blind” Image Quality Analyzer}.
\newblock \bibinfo{journal}{\emph{IEEE Signal Processing Letters}} \bibinfo{volume}{20}, \bibinfo{number}{3} (\bibinfo{year}{2013}), \bibinfo{pages}{209--212}.
\newblock
\href{https://doi.org/10.1109/LSP.2012.2227726}{doi:\nolinkurl{10.1109/LSP.2012.2227726}}


\bibitem[Monakhova et~al\mbox{.}(2022)]%
        {monakhova2022starlight}
\bibfield{author}{\bibinfo{person}{Kristina Monakhova}, \bibinfo{person}{Stephan~R. Richter}, \bibinfo{person}{Laura Waller}, {and} \bibinfo{person}{Vladlen Koltun}.} \bibinfo{year}{2022}\natexlab{}.
\newblock \showarticletitle{Dancing Under the Stars: Video Denoising in Starlight}. In \bibinfo{booktitle}{\emph{Proceedings of the IEEE/CVF Conference on Computer Vision and Pattern Recognition (CVPR)}}. \bibinfo{pages}{16241--16251}.
\newblock


\bibitem[Perazzi et~al\mbox{.}(2016)]%
        {Perazzi2016davis}
\bibfield{author}{\bibinfo{person}{F. Perazzi}, \bibinfo{person}{J. Pont-Tuset}, \bibinfo{person}{B. McWilliams}, \bibinfo{person}{L. {Van Gool}}, \bibinfo{person}{M. Gross}, {and} \bibinfo{person}{A. Sorkine-Hornung}.} \bibinfo{year}{2016}\natexlab{}.
\newblock \showarticletitle{A Benchmark Dataset and Evaluation Methodology for Video Object Segmentation}. In \bibinfo{booktitle}{\emph{Computer Vision and Pattern Recognition}}.
\newblock


\bibitem[Ronneberger et~al\mbox{.}(2015)]%
        {ronneberge2015unet}
\bibfield{author}{\bibinfo{person}{Olaf Ronneberger}, \bibinfo{person}{Philipp Fischer}, {and} \bibinfo{person}{Thomas Brox}.} \bibinfo{year}{2015}\natexlab{}.
\newblock \showarticletitle{U-Net: Convolutional Networks for Biomedical Image Segmentation}. In \bibinfo{booktitle}{\emph{Medical Image Computing and Computer-Assisted Intervention -- MICCAI 2015}}, \bibfield{editor}{\bibinfo{person}{Nassir Navab}, \bibinfo{person}{Joachim Hornegger}, \bibinfo{person}{William~M. Wells}, {and} \bibinfo{person}{Alejandro~F. Frangi}} (Eds.). \bibinfo{publisher}{Springer International Publishing}, \bibinfo{address}{Cham}, \bibinfo{pages}{234--241}.
\newblock
\showISBNx{978-3-319-24574-4}


\bibitem[Rosenblatt(1958)]%
        {rosenblatt1958perceptron}
\bibfield{author}{\bibinfo{person}{F. Rosenblatt}.} \bibinfo{year}{1958}\natexlab{}.
\newblock \showarticletitle{The Perceptron: A Probabilistic Model for Information Storage and Organization in the Brain}.
\newblock \bibinfo{journal}{\emph{Psychological Review}} \bibinfo{volume}{65}, \bibinfo{number}{6} (\bibinfo{year}{1958}), \bibinfo{pages}{386--408}.
\newblock
\href{https://doi.org/10.1037/h0042519}{doi:\nolinkurl{10.1037/h0042519}}


\bibitem[Sigurdsson et~al\mbox{.}(2016)]%
        {sigurdsson2016charades}
\bibfield{author}{\bibinfo{person}{Gunnar~A. Sigurdsson}, \bibinfo{person}{G{\"u}l Varol}, \bibinfo{person}{Xiaolong Wang}, \bibinfo{person}{Ivan Laptev}, \bibinfo{person}{Ali Farhadi}, {and} \bibinfo{person}{Abhinav Gupta}.} \bibinfo{year}{2016}\natexlab{}.
\newblock \showarticletitle{Hollywood in Homes: Crowdsourcing Data Collection for Activity Understanding}.
\newblock \bibinfo{journal}{\emph{ArXiv e-prints}} (\bibinfo{year}{2016}).
\newblock
\showeprint{1604.01753}
\urldef\tempurl%
\url{http://arxiv.org/abs/1604.01753}
\showURL{%
\tempurl}


\bibitem[Wang et~al\mbox{.}(2023)]%
        {wang2022clipiqa}
\bibfield{author}{\bibinfo{person}{Jianyi Wang}, \bibinfo{person}{Kelvin~CK Chan}, {and} \bibinfo{person}{Chen~Change Loy}.} \bibinfo{year}{2023}\natexlab{}.
\newblock \showarticletitle{Exploring CLIP for Assessing the Look and Feel of Images}. In \bibinfo{booktitle}{\emph{AAAI}}.
\newblock


\bibitem[Wang et~al\mbox{.}(2021)]%
        {wang2021sdsd}
\bibfield{author}{\bibinfo{person}{Ruixing Wang}, \bibinfo{person}{Xiaogang Xu}, \bibinfo{person}{Chi-Wing Fu}, \bibinfo{person}{Jiangbo Lu}, \bibinfo{person}{Bei Yu}, {and} \bibinfo{person}{Jiaya Jia}.} \bibinfo{year}{2021}\natexlab{}.
\newblock \showarticletitle{Seeing Dynamic Scene in the Dark: High-Quality Video Dataset with Mechatronic Alignment}. In \bibinfo{booktitle}{\emph{ICCV}}.
\newblock


\bibitem[Wang et~al\mbox{.}(2024)]%
        {wang2024lmot}
\bibfield{author}{\bibinfo{person}{Xinzhe Wang}, \bibinfo{person}{Kang Ma}, \bibinfo{person}{Qiankun Liu}, \bibinfo{person}{Yunhao Zou}, {and} \bibinfo{person}{Ying Fu}.} \bibinfo{year}{2024}\natexlab{}.
\newblock \showarticletitle{Multi-Object Tracking in the Dark}. In \bibinfo{booktitle}{\emph{Proceedings of the IEEE/CVF Conference on Computer Vision and Pattern Recognition (CVPR)}}. \bibinfo{pages}{382--392}.
\newblock


\bibitem[Wei et~al\mbox{.}(2021)]%
        {wei2021physicsmodeling}
\bibfield{author}{\bibinfo{person}{Kaixuan Wei}, \bibinfo{person}{Ying Fu}, \bibinfo{person}{Yinqiang Zheng}, {and} \bibinfo{person}{Jiaolong Yang}.} \bibinfo{year}{2021}\natexlab{}.
\newblock \showarticletitle{Physics-based noise modeling for extreme low-light photography}.
\newblock \bibinfo{journal}{\emph{IEEE Transactions on Pattern Analysis and Machine Intelligence}} \bibinfo{volume}{44}, \bibinfo{number}{11} (\bibinfo{year}{2021}), \bibinfo{pages}{8520--8537}.
\newblock


\bibitem[Xu et~al\mbox{.}(2018)]%
        {xu2018ytvos}
\bibfield{author}{\bibinfo{person}{Ning Xu}, \bibinfo{person}{Linjie Yang}, \bibinfo{person}{Yuchen Fan}, \bibinfo{person}{Dingcheng Yue}, \bibinfo{person}{Yuchen Liang}, \bibinfo{person}{Jianchao Yang}, {and} \bibinfo{person}{Thomas~S. Huang}.} \bibinfo{year}{2018}\natexlab{}.
\newblock \showarticletitle{YouTube-VOS: {A} Large-Scale Video Object Segmentation Benchmark}.
\newblock \bibinfo{journal}{\emph{CoRR}}  \bibinfo{volume}{abs/1809.03327} (\bibinfo{year}{2018}).
\newblock
\urldef\tempurl%
\url{http://arxiv.org/abs/1809.03327}
\showURL{%
\tempurl}


\bibitem[Yang et~al\mbox{.}(2019)]%
        {yang2019ytvis}
\bibfield{author}{\bibinfo{person}{Linjie Yang}, \bibinfo{person}{Yuchen Fan}, {and} \bibinfo{person}{Ning Xu}.} \bibinfo{year}{2019}\natexlab{}.
\newblock \showarticletitle{Video instance segmentation}. In \bibinfo{booktitle}{\emph{ICCV}}.
\newblock


\bibitem[Ye et~al\mbox{.}(2022)]%
        {ye2022darktrack}
\bibfield{author}{\bibinfo{person}{Junjie Ye}, \bibinfo{person}{Changhong Fu}, \bibinfo{person}{Ziang Cao}, \bibinfo{person}{Shan An}, \bibinfo{person}{Guangze Zheng}, {and} \bibinfo{person}{Bowen Li}.} \bibinfo{year}{2022}\natexlab{}.
\newblock \showarticletitle{{Tracker Meets Night: A Transformer Enhancer for UAV Tracking}}.
\newblock \bibinfo{journal}{\emph{IEEE Robotics and Automation Letters}} \bibinfo{volume}{7}, \bibinfo{number}{2} (\bibinfo{year}{2022}), \bibinfo{pages}{3866--3873}.
\newblock
\href{https://doi.org/10.1109/LRA.2022.3146911}{doi:\nolinkurl{10.1109/LRA.2022.3146911}}


\bibitem[Yi and Anantrasirichai(2024)]%
        {Yi2024trackinglowlight}
\bibfield{author}{\bibinfo{person}{Anqi Yi} {and} \bibinfo{person}{Nantheera Anantrasirichai}.} \bibinfo{year}{2024}\natexlab{}.
\newblock \showarticletitle{A Comprehensive Study of Object Tracking in Low-Light Environments}.
\newblock \bibinfo{journal}{\emph{arXiv:2312.16250}} (\bibinfo{year}{2024}).
\newblock


\bibitem[Zamir et~al\mbox{.}(2020)]%
        {zamir2020cycleisp}
\bibfield{author}{\bibinfo{person}{Syed~Waqas Zamir}, \bibinfo{person}{Aditya Arora}, \bibinfo{person}{Salman Khan}, \bibinfo{person}{Munawar Hayat}, \bibinfo{person}{Fahad~Shahbaz Khan}, \bibinfo{person}{Ming-Hsuan Yang}, {and} \bibinfo{person}{Ling Shao}.} \bibinfo{year}{2020}\natexlab{}.
\newblock \showarticletitle{CycleISP: Real Image Restoration via Improved Data Synthesis}. In \bibinfo{booktitle}{\emph{Proceedings of the IEEE/CVF Conference on Computer Vision and Pattern Recognition (CVPR)}}.
\newblock


\bibitem[Zhang et~al\mbox{.}(2023)]%
        {Zhang2023RAWnoise}
\bibfield{author}{\bibinfo{person}{Feng Zhang}, \bibinfo{person}{Bin Xu}, \bibinfo{person}{Zhiqiang Li}, \bibinfo{person}{Xinran Liu}, \bibinfo{person}{Qingbo Lu}, \bibinfo{person}{Changxin Gao}, {and} \bibinfo{person}{Nong Sang}.} \bibinfo{year}{2023}\natexlab{}.
\newblock \showarticletitle{Towards General Low-Light Raw Noise Synthesis and Modeling}. In \bibinfo{booktitle}{\emph{Proceedings of the IEEE/CVF International Conference on Computer Vision (ICCV)}}. \bibinfo{pages}{10820--10830}.
\newblock


\bibitem[Zhang et~al\mbox{.}(2018)]%
        {zhang2018lpips}
\bibfield{author}{\bibinfo{person}{Richard Zhang}, \bibinfo{person}{Phillip Isola}, \bibinfo{person}{Alexei~A Efros}, \bibinfo{person}{Eli Shechtman}, {and} \bibinfo{person}{Oliver Wang}.} \bibinfo{year}{2018}\natexlab{}.
\newblock \showarticletitle{The Unreasonable Effectiveness of Deep Features as a Perceptual Metric}. In \bibinfo{booktitle}{\emph{CVPR}}.
\newblock


\bibitem[Zhou et~al\mbox{.}(2022)]%
        {zhou2022lednet}
\bibfield{author}{\bibinfo{person}{Shangchen Zhou}, \bibinfo{person}{Chongyi Li}, {and} \bibinfo{person}{Chen~Change Loy}.} \bibinfo{year}{2022}\natexlab{}.
\newblock \showarticletitle{LEDNet: Joint Low-light Enhancement and Deblurring in the Dark}. In \bibinfo{booktitle}{\emph{ECCV}}.
\newblock


\end{thebibliography}

\end{document}